\documentclass[10pt,twocolumn,letterpaper]{article}

\newif\ifreview 
\newif\ifarxiv \newcommand{\arxiv}{\arxivtrue}
\newif\ifcamera 

\arxiv

\ifreview \usepackage[review]{cvpr} \fi     % To produce the REVIEW version
\ifcamera \usepackage{cvpr} \fi             % To produce the CAMERA-READY version
\ifarxiv \usepackage[pagenumbers]{cvpr} \fi % To force page numbers, e.g. for an arXiv version

\makeatletter
\@namedef{ver@everyshi.sty}{}
\makeatother

\usepackage{graphicx}
\usepackage{caption}
\usepackage{amsmath,bm}
\usepackage{amssymb}
\usepackage{booktabs}
\usepackage{svg}
\usepackage{multirow}

\newcommand{\bv}[1]{{\bm {#1}}}

\usepackage[pagebackref,breaklinks,colorlinks]{hyperref}

\usepackage[capitalize]{cleveref}
\crefname{section}{Sec.}{Secs.}
\Crefname{section}{Section}{Sections}
\Crefname{table}{Table}{Tables}
\crefname{table}{Tab.}{Tabs.}

\def\cvprPaperID{10479} % *** Enter the CVPR Paper ID here
\def\confName{CVPR}
\def\confYear{2023}

\begin{document}

\newcommand{\M}{FeatureBooster}

\title{\M: Boosting Feature Descriptors with a Lightweight Neural Network}

\author{Xinjiang Wang\textsuperscript{1,2} \qquad
Zeyu Liu\textsuperscript{1,2} \qquad
Yu Hu\textsuperscript{1,2} \qquad
Wei Xi\textsuperscript{3} \qquad
Wenxian Yu\textsuperscript{1,2} \qquad
Danping Zou\textsuperscript{1,2}\thanks{ Corresponding Author: Danping Zou ({\tt\small dpzou@sjtu.edu.cn}). This works was supported by National Key R\&D Program (2022YFB3903802) and National of Science Foundation of China (62073214)}. \qquad
\\
\textsuperscript{1}{Shanghai Key Laboratory of Navigation and Location Based Services, Shanghai Jiao Tong University} \\
\textsuperscript{2}{SJTU SEIEE $\cdot$ G60 Yun Zhi AI Innovation and Application Research Center} \\
\textsuperscript{3}{Intelligent Perception Institute, Midea Corporate Research Center} \\
\tt\small {\{wangxj83,ribosomal,henryhuyu,wxyu,dpzou\}@sjtu.edu.cn} \qquad
\tt\small {xiwei1@midea.com}
}

%\thanks{This work was supported by NSFC(62073214).}
% \author{Xinjiang Wang\\
% Shanghai Jiao Tong University\\
% Shanghai, China\\
% {\tt\small wangxj83@sjtu.edu.cn}
% % \and
% % Second Author\\
% % First line of institution2 address\\
% % {\tt\small secondauthor@i2.org}
% }

\maketitle

%%%%%%%%% ABSTRACT
\begin{abstract}
  We introduce a lightweight network to improve descriptors of keypoints within the same image. The network takes the original descriptors and the geometric properties of keypoints as the input, and uses an MLP-based self-boosting stage and a Transformer-based cross-boosting stage to enhance the descriptors. The boosted descriptors can be either real-valued or binary ones. We use the proposed network to boost both hand-crafted (ORB\cite{orb}, SIFT\cite{sift}) and the state-of-the-art learning-based descriptors (SuperPoint\cite{sp}, ALIKE\cite{alike}) and evaluate them on image matching, visual localization, and structure-from-motion tasks. The results show that our method significantly improves the performance of each task, particularly in challenging cases such as large illumination changes or repetitive patterns. Our method requires only 3.2ms on desktop GPU and 27ms on embedded GPU to process 2000 features, which is fast enough to be applied to a practical system. 
  The code and trained weights are publicly available at \href{https://github.com/SJTU-ViSYS/FeatureBooster}{github.com/SJTU-ViSYS/FeatureBooster}.
\end{abstract}

%%%%%%%%% BODY TEXT
%-------------------Introduction--------------------
\section{Introduction}
\label{sec:intro}
Extracting sparse keypoints or local features from an image is a fundamental building block in various computer vision tasks, such as structure from motion (SfM), simultaneous localization and mapping (SLAM), and visual localization. The feature descriptor, represented by a real-valued or binary descriptor,  
plays a key role in matching those keypoints across different images. 

The descriptors are commonly hand-crafted in the early days. Recently, learning-based descriptors \cite{sp,alike} have shown to be more powerful than hand-crafted ones, especially in challenging cases such as significant viewpoint and illumination changes. Both hand-crafted and learning-based descriptors have shown to work well in practice. Some of them have become default descriptors for some applications. For example, the simple binary descriptor ORB\cite{orb} is widely used for SLAM systems\cite{orbslam2,okvis}. SIFT\cite{sift} is typically used in structure-from-motion systems. 

Considering that the descriptors have already been integrated into practical systems, replacing them with totally new ones can be problematic, as it may require more computing power that may not be supported by the existing hardware, or sometimes require extensive modifications to the software because of changed descriptor type (\eg from binary to real). 

\begin{figure}[t]
  \centering
%   \fbox{\rule{0pt}{2in} \rule{0.9\linewidth}{0pt}}
  \includegraphics[width=1.0\linewidth]{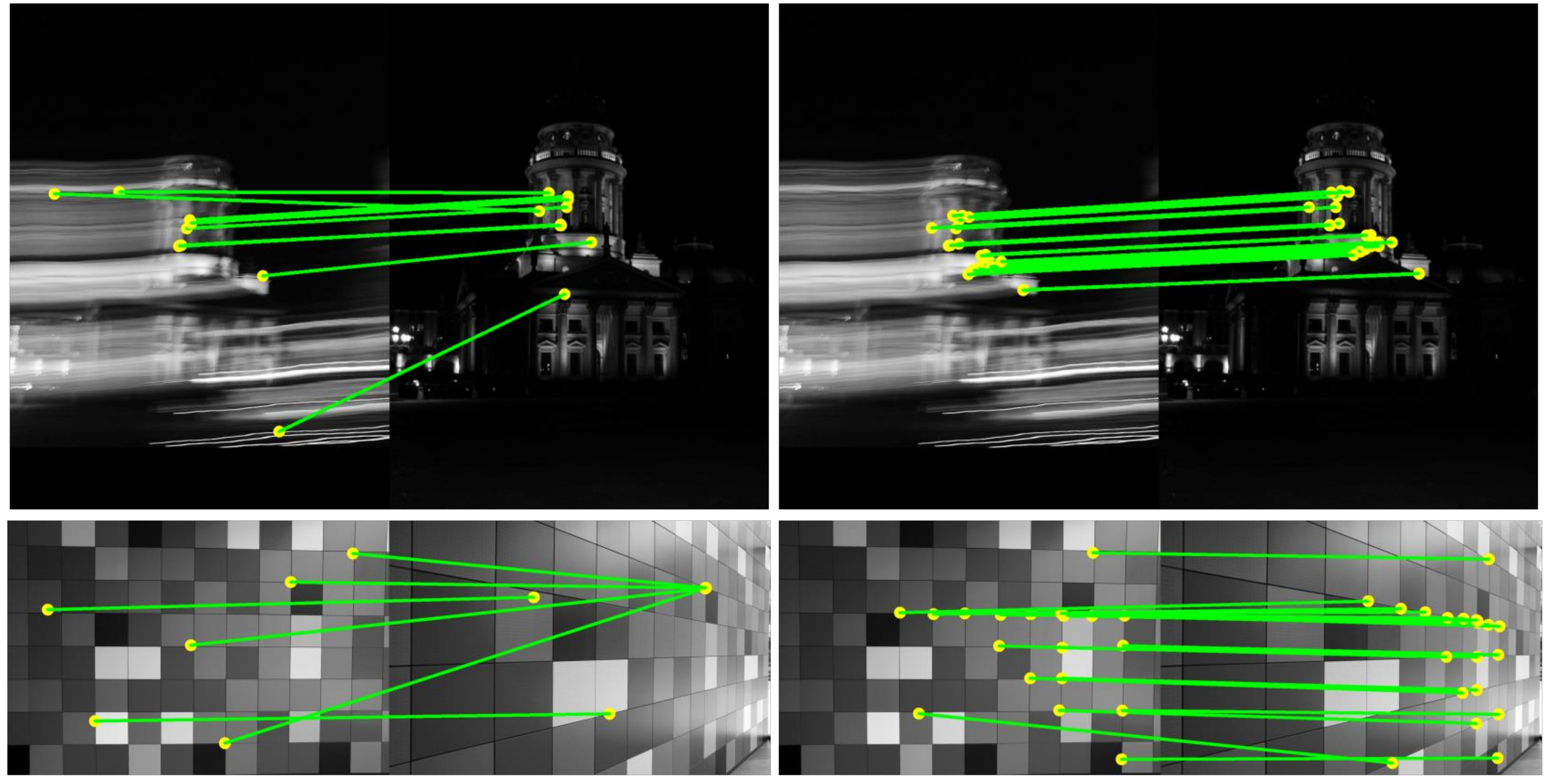}
  \caption{ORB descriptors perform remarkably better in challenging cases after being boosted by the proposed lightweight network. \textbf{Left column}: Matching results of using raw ORB descriptors. \textbf{Right column}: Results of using boosted ORB descriptors. Nearest neighbor search and RANSAC\cite{ransac} were used for matching.}
  \label{fig:matches}
\end{figure}

In this work, we attempt to reuse existing descriptors and enhance their discrimination ability with as little computational overhead as possible. To this end, we propose a lightweight network to improve the original descriptors. The input of this network is the descriptors and the geometric properties such as the 2D locations of all the keypoints within the entire image. Each descriptor is firstly processed by an MLP (Multi-layer perceptron) and summed with geometric properties encoded by another MLP. The new geometrically encoded descriptors are then aggregated by an efficient Transformer to produce powerful descriptors that are aware of the high-level visual context and spatial layout of those keypoints. The enhanced descriptors can be either real-valued or binary ones and matched by using Euclidean/Hamming distance respectively.

The core idea of our approach, motivated by recent work \cite{sg,loftr, contextdesc}, is integrating the visual and geometric information of all the keypoints into individual descriptors by a Transformer. This can be better understood intuitively by considering when people are asked to find correspondences between images, they would check all the keypoints and the spatial layout of those keypoints in each image. With the help of the global receptive field in Transformer, the boosted descriptors contain global contextual information that makes them more robust and discriminative as shown in \cref{fig:matches}. 

We apply our \M~to both hand-crafted descriptors (SIFT\cite{sift}, ORB\cite{orb}) and the state-of-the-art learning-based descriptors (SuperPoint\cite{sp}, ALIKE\cite{alike}). We evaluated the boosted descriptors on tasks including image matching, visual localization, and structure-from-motion. The results show that our method can significantly improve the performance of each task by using our boosted descriptors. 

Because \M~does not need to process the image and adopts a lightweight Transformer, it is highly efficient. It takes only 3.2ms on NVIDIA RTX 3090 and 27ms on NVIDIA Jetson Xavier NX (for embedded devices) to boost 2000 features, which makes our method applicable to practical systems. 

% Besides, we also demonstrate the effectiveness and robustness of our work in visual SLAM by integrate \M
%  into ORB-SLAM2\cite{orbslam2}, which can run real-time on embedded low power hardware, such as a Jetson Xaiver NX.
 
%-------------------Related work--------------------
\section{Related work}
\label{sec:related}

\noindent
\textbf{Feature descriptors:} 
%Feature descriptors can be roughly divided into two major categories: hand-crafted descriptors and learning-based descriptors.
For a long time, the descriptors are commonly hand-crafted. SIFT\cite{sift} and ORB\cite{orb} are the most well-known hand-crafted descriptors, which are still widely used in many 3D computer vision tasks for their good performance and high efficiency. Hand-crafted descriptors are usually extracted from a local patch. It hence limits their representation capability on higher levels. 
With the development of deep learning and the emergence of patch dataset with annotation\cite{brown}, learning-based descriptors have been widely studied. Most learning-based descriptors from patches adopt the network architecture introduced in L2-Net\cite{l2net} and are trained with different loss functions, \eg triplet loss\cite{hardnet,sosnet,hynet}, N-Pair loss\cite{l2net} and list-wise ranking loss\cite{doap}. 
Learning-based dense descriptors\cite{lfnet,sp,d2net,r2d2,s2dnet,caps} can leverage information beyond local patches in that they are typically extracted from the entire image using convolutional neural networks, thus exhibiting superior performances on large viewpoint and illumination changes.
Though a lot of descriptors have been invented, how to boost existing descriptors has received little attention, particularly through a learning-based approach.

\noindent
\textbf{Improve existing feature descriptors:} 
It has been found that projecting existing descriptors into another space by a non-linear transformation leads to better matching results\cite{descriptorlearning}. RootSIFT \cite{rootsift} shows that simply taking the square root of each element of the normalized SIFT descriptors can improve the matching results. Apart from improving the discrimination, some works also seek to compress the descriptors by reducing the descriptor's dimension, such as PCA-SIFT\cite{pcasift} and LDAHash\cite{ldahash}.  A recent work  \cite{crossdr} trained a network to map different types of descriptors into a common space such that different types of descriptors can be matched. Our work shares the core idea with this line of research but aims to enhance the discrimination ability to exist descriptors using a lightweight neural network. 

\noindent
\textbf{Feature matching:} 
Once feature descriptors are acquired, the correspondences between images are usually found by nearest neighbor (NN) search. The incorrect matches can be filtered by adopting some tricks (\eg mutual check, Lowe's ratio test\cite{sift}, and RANSAC\cite{ransac}). 
However, NN search ignores the spatial and visual relationship between features and usually produces noisy matching results. To address this problem, SuperGlue\cite{sg} trained an attentional graph neural network by correlating two sets of local features from different images to predict the correspondences. Our approach is largely inspired by SuperGlue, but does not attempt to improve the matching process. It instead enhances the feature descriptors from a single image, such that a simple NN search can be used to produce competitive results. Therefore our approach can be seamlessly integrated into many existing pipelines such as a BoW(bag-of-word)\cite{dbow} implementation.

\noindent
\textbf{Feature context:}
The distribution of feature locations and descriptors within an entire image forms a global context that can be helpful for feature matching as demonstrated in SuperGlue\cite{sg}. In this paper, we aim to integrate the global context information into original descriptors to boost their discrimination ability rather than learning to describe the image from scratch.
The closest work to our approach is SConE\cite{scone} and ContextDesc\cite{contextdesc}. 
SConE\cite{scone} develops a constellation embedding module to convert a set of adjacent features (including original descriptors and their spatial layout) into new descriptors. This module is designed for a particular type of descriptor (FREAK\cite{freak}).
ContextDesc\cite{contextdesc} uses two MLPs to encode the visual context and geometric context into global features to improve the local descriptors. It however requires to use of extra CNN to extract high-level features from the original image to construct the visual context.

By contrast, our method takes only the descriptors and geometric information (such as 2D locations) as the input and uses a lightweight Transformer to aggregate them to produce new descriptors. The new descriptors can be both binary or real-valued ones and can be seamlessly integrated into existing visual localization, SLAM, and structure-from-motion systems. No need to process the raw images makes our method very efficient and can run in real-time on embedded GPU devices. 

%-------------------Method--------------------------
\section{Overview}
\label{sec:method}
\begin{figure*}[t]
  \centering
%   \fbox{\rule{0pt}{2in} \rule{.9\linewidth}{0pt}}
  \includegraphics[width=0.85\linewidth]{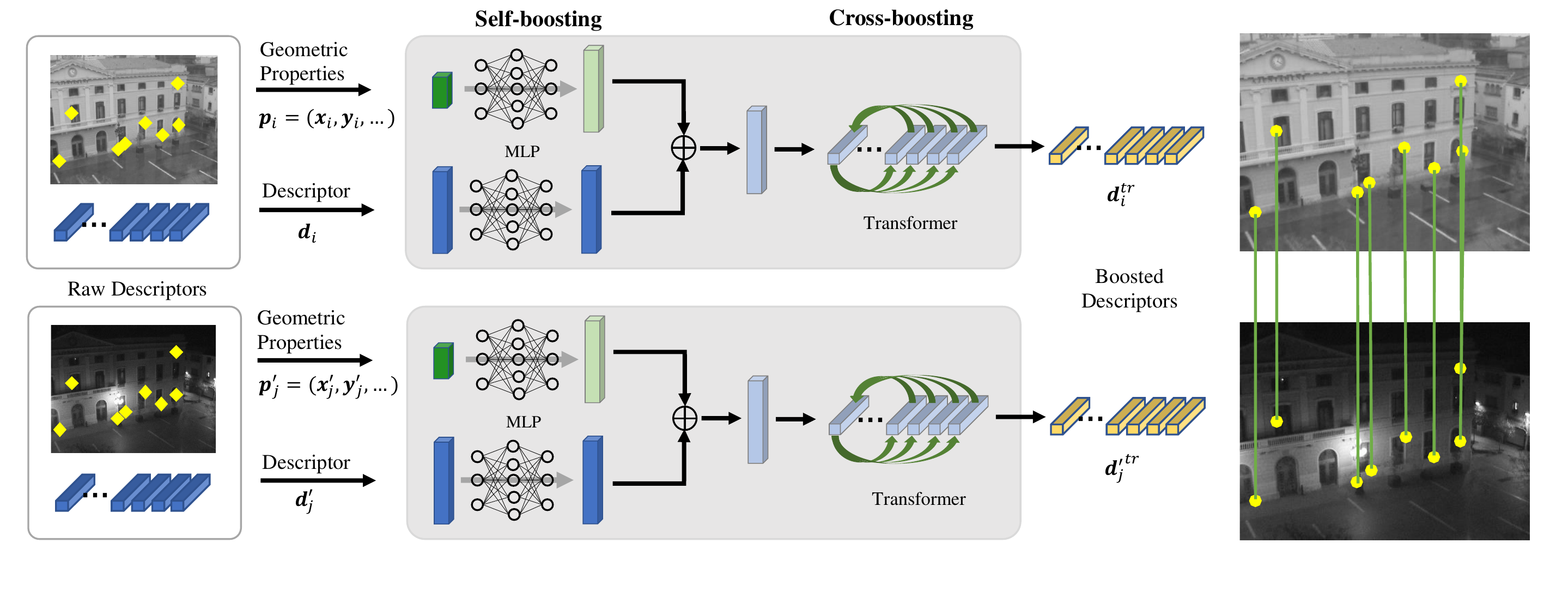}
  \vspace{-1em}
  \caption{The proposed \emph{FeatureBooster} pipeline consists of self-boosting and cross-boosting stages. Self-boosting applies an MLP to encode the geometric properties of a keypoint and combines it with a new descriptor projected by another MLP. In the cross-boosting stage, the geometrically encoded descriptors of all the keypoints within the entire image are then sent to a lightweight Transformer to generate boosted descriptors. Finally, the boosted descriptors are used for feature matching.}
  \label{fig:pipline}
\end{figure*}
We propose a lightweight network to boost the feature vectors (or descriptors) of a set of keypoints extracted from an image by some existing keypoint detectors as shown in \cref{fig:pipline}. It takes only the feature descriptors as well as the geometric information such as feature position, orientation, and scale as the input, and outputs new descriptors that are much more powerful than the original ones. The new descriptors can be either real-valued or binary vectors which may be different from the original ones. 
Our feature booster does not need to process the image from which those keypoints are extracted, which makes our model lightweight and efficient, and can be more easily integrated into existing Structure-from-motion or SLAM systems. No need to access the original images also makes our approach possible to reuse 3D maps already built with certain types of features.

The proposed pipeline consists of two steps: \textbf{Self-boosting} and \textbf{Cross-boosting}. Self-boosting refers to using a lightweight MLP network to project the original feature vector into a new space. It also encodes geometric information such as  2D location, detection score, and orientation/scale to a high-dimensional vector to improve the descriptor. After that, cross-boosting explores the global context including the descriptors of other features and the spatial layout of all the features to further enhance the individual descriptors using a lightweight Transformer. 
The proposed network is trained end-to-end by using a loss function that consists of a ranking-based retrieval loss and an enhancement loss.
% Finally, we introduce the implementation details of our method.

%--------------------------------
\subsection{Self-boosting}
For each keypoint $i$ detected in the image, we can obtain its visual descriptor $\bv{d}_i$, a $D$ dimensional real-valued or binary vector. The feature descriptors are then used to establish the correspondences between images by measuring their similarity. A powerful descriptor should be robust to the viewpoint and illumination changes to produce correct matching results. A lot of descriptors have been developed, including hand-crafted methods such as ORB \cite{orb}, SURF \cite{surf}, and SIFT \cite{sift}, as well as more advanced learning-based methods such as SuperPoint \cite{sp}. However, there are still some problems with those descriptors. 

For the hand-crafted ones, the first problem is that the similarity metric in the descriptor space is not optimal for feature matching. This has been noticed in \cite{rootsift}, where a Hellinger distance is used to measure the SIFT's similarity instead using a Euclidean distance,  which leads to a better matching performance.  It can be seen from \cite{descriptorlearning}, changing the similarity metric is equivalent to projecting the original descriptors into another space. This motivates us to use an MLP (Multi-layer perceptron) to map the original descriptor into a new one.

MLP is a universal function approximator as shown by Cybenko's theorem\cite{cybenko}. Hence we can use an MLP to approximate the project function which we refer to as $\mathbf{MLP}_{desc}$. 
The transformed descriptor $\bv{d}^{tr}_i$ for keypoint $i$ is the non-linear projection of the extracted descriptor $\bv{d}_i$:
\begin{equation}
  \bv{d}^{tr}_i \leftarrow \mathbf{MLP}_{desc}(\bv{d}_i)
  \label{eq:proj_mlp}
\end{equation}

Given that the network's training phase is guided by a loss function with Euclidean or Hamming distance constraints, this MLP-based model enables the transformed descriptors to be well fit for measuring similarity in Euclidean or Hamming space respectively, especially for the hand-crafted descriptors.
However, this projection hasn't exploited the geometric information of the key point which is valuable for matching \cite{sg}. Therefore, we also embed the geometric information into a high dimensional vector using another MLP ($\mathbf{MLP}_{geo}) $ to further improve the descriptor. 
We encode not only the 2D location of keypoints $(x_i, y_i)$, but also other information such as the scale $s_i$, orientation $\theta_i$, and detection score $c_i$ when they are available. The high-dimensional embedded geometric information is added to the transformed descriptor:
\begin{equation}
  \bv{d}^{tr}_i \leftarrow \bv{d}^{tr}_i + \mathbf{MLP}_{geo}(\bv{p}_i).
  \label{eq:kenc}
\end{equation}
Here, $\bv{p}_i = (x_i,y_i,c_i,\theta_i, s_i)$ represents all available geometric information as aforementioned.
% \begin{figure}[t]
%   \centering
% %   \fbox{\rule{0pt}{2in} \rule{0.9\linewidth}{0pt}}
%   \includegraphics[width=0.6\linewidth]{figure/refinenet-mlp.pdf}
%   \caption{MLP-based projection.}
%   \label{fig:mlp}
% \end{figure}

%--------------------------------
\subsection{Cross-boosting}
Self-boosting enhances the descriptor of each keypoint independently without considering the possible correlation between different keypoints. For example, it does not exploit the spatial relationships between those keypoints, while the spatial contextual cues could greatly enhance the matching capability as demonstrated in \cite{sg}. Therefore, the boosted descriptors from the self-boosting stage are limited to the local context and still perform poorly under some challenging environments (\eg repetitive patterns or weakly textured scenes). To address this issue, we further process those descriptors by a cross-boosting stage. 

Motivated by SuperGlue\cite{sg}, we use a Transformer to capture spatial contextual cues of the sparse local features extracted from the same image.  We denote the Transformer by $\mathbf{Trans}$ and the projection is described as:
\begin{equation}
  (\bv{d}^{tr}_1,\bv{d}^{tr}_2,\ldots \bv{d}^{tr}_N) \leftarrow \mathbf{Trans}(\bv{d}^{tr}_1,\bv{d}^{tr}_2,\ldots \bv{d}^{tr}_N),
  \label{eq:proj_t}
\end{equation}
where the input of the Transformer is $N$ local features within the same image, and the output is the enhanced feature descriptors.
Compared with the MLP-based projection (see \cref{eq:proj_mlp}), Transformer-based projection processes all the local features within the same image simultaneously.
With the help of the attention mechanism in Transformer, all local features' information can be aggregated to form a global context.
By integrating this global contextual information, the local feature descriptors may have larger receptive fields and adjust themselves according to their neighbors (or competitors in the case of feature matching). Therefore their distinguishability can be improved, especially for local features extracted from repetitive patterns as shown in \cref{fig:matches}.

\begin{figure}[t]
  \centering
%   \fbox{\rule{0pt}{2in} \rule{0.9\linewidth}{0pt}}
  \includegraphics[width=0.8\linewidth]{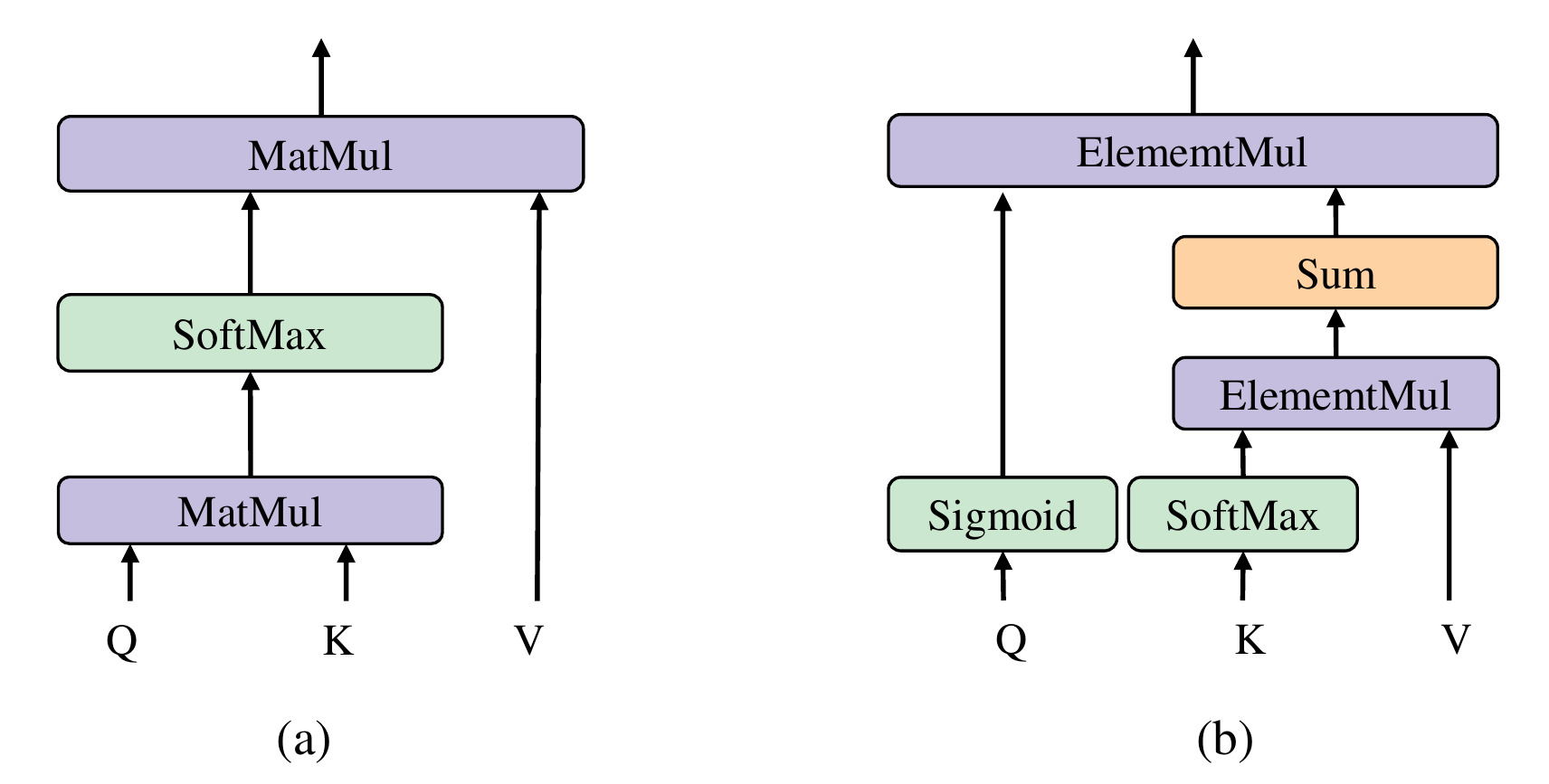}
  \caption{Different architectures of the attention layer. (a) Attention layer in a vanilla Transformer. (b) Attention-Free Transformer (AFT-simple), where only element-wise multiplication is required.}
  \label{fig:atten}
\end{figure}

The biggest issue of using a Transformer is that its attention mechanism requires high memory and computation costs.
The transformer encoder layer consists of two sublayers: an attention layer and a position-wise fully connected feed-forward network. The vanilla Transformer\cite{transformer} uses a Multi-Head Attention (MHA) layer. Given an input $\mathbf{X}\in\mathbb{R}^{N\times D}$, where the $i$-th row is the $D$ dimensional feature vector of keypoint $i$, the $h$-th head attention of $\mathbf{X}$ is defined as:
\begin{equation}
\small
  \begin{split}
      \mathbf{f}_h(\mathbf{X}) & = \mathit{softmax}(\frac{\mathbf{Q}_h\mathbf{K}_h^\top}{D_k})\mathbf{V}_h, \\
      & s.t.\ \mathbf{Q}_h = \mathbf{X}\mathbf{W}_h^Q, \mathbf{K}_h = \mathbf{X}\mathbf{W}_h^K, \mathbf{V}_h = \mathbf{X}\mathbf{W}_h^V
  \end{split}
  \label{eq:attention}
\end{equation}
where $\mathbf{W}_h^Q\in\mathbb{R}^{D\times D_k}, \mathbf{W}_h^K\in\mathbb{R}^{D\times D_k}, \mathbf{W}_h^V\in\mathbb{R}^{D\times D_v}$ are the linear projections of for head $h$. 
\cref{fig:atten}(a) illustrates the computation graph of dot-product attention. 
The output of Multi-Head Attention is the concatenation of all the attention heads' outputs along the channel dimension.

MHA uses the attention matrix to enable the global interaction between query and value. 
The computation of the attention matrix relies on the matrix dot product between query and key, which results in a time and space complexity quadratic with the context size ($O(N^2D)$).
It is easy to see that the complexity introduced by MHA makes Vanilla Transformer difficult to scale to inputs with a large context size ($N$).
In our case, the context size ($N$) is the number of local features within an image. Unfortunately, it is very common that thousands of local features have been extracted within one image.

%\begin{figure}[!t]
%  \centering
%   \fbox{\rule{0pt}{2in} \rule{0.9\linewidth}{0pt}}
%  \includegraphics[width=0.7\linewidth]{figure/refinenet-transformer.pdf}
%  \caption{Transformer-based projection.}
%  \label{fig:trans}
%\end{figure}
\textbf{Attention-Free Transformer:}
To address the scalability problem in our case, we propose to use an efficient Attention-Free Transformer (specifically AFT-Simple) \cite{aft} to replace the MHA operation in a Vanilla Transformer. 
Unlike MHA or recent linearized attention\cite{la}, Attention-Free Transformer (AFT) does not use or approximate the dot product attention. 
Specifically, AFT rearranges the computation order of Q, K, and V, just like linear attention, but multiplies K and V element-wise instead of using matrix multiplication. The Attention-Free Transformer for keypoint $i$ can be formulated as:
\begin{equation}
\small
  \begin{split}
  \mathbf{f}_i(\mathbf{X}) & = \sigma(\mathbf{Q}_{i})\odot\frac{\sum^N_{j=1}\exp(\mathbf{K}_{j})\odot\mathbf{V}_{j}}{\sum^N_{j=1}\exp(\mathbf{K}_{j})} \\
  & = \sigma(\mathbf{Q}_{i})\odot\sum^N_{j=1}(\mathit{softmax}(\mathbf{K})\odot\mathbf{V})_{j}
  \end{split}
  \label{eq:aft}
\end{equation}
where $\sigma(\cdot)$ is a Sigmoid function;  $\mathbf{Q}_i$ represents $i$-th row of $\mathbf{Q}$; $\mathbf{K}_j,\mathbf{V}_j$ represent the $j$-th rows of $\mathbf{K},\mathbf{V}$.
AFT-simple performs a revised version of the MHA operation where the number of attention heads is equal to the model’s feature dimension $D$ and the similarity used in MHA is replaced by a kernel function $sim(\mathbf{Q},\mathbf{K})=\sigma(\mathbf{Q})\cdot\mathit{softmax}(\mathbf{K})$.
In this way, attention can be computed by element-wise multiplication instead of matrix multiplication, which results in a time and space complexity that is linear with context and feature size ($O(ND)$).
\cref{fig:atten}(b) illustrates the computation graph of AFT-Simple. 

%--------------------------------
\subsection{Loss Functions}
As in previous work\cite{doap,r2d2}, we treat the descriptor matching problem as nearest neighbor retrieval and use the Average Precision (AP) to train the descriptors. Considering transformed local feature descriptors $\mathbf{d^{tr}} = (\bv{d}_1^{tr},\ldots,\bv{d}_N^{tr})$, we want to maximize the AP\cite{ap} for all descriptors and our goal for training is to minimize the following cost function:
\begin{equation}
  \mathcal{L}_{AP} = 1 - \frac{1}{N}(\sum^N_{i}\mathit{AP}(\bv{d}^{tr}_{i}))
  \label{eq:aploss}
\end{equation}
To ensure that the original descriptors will be boosted, we propose to use another loss to force the performance of transformed descriptors to be better than the original ones:
\begin{equation}
  \mathcal{L}_{BOOST} = \frac{1}{N}\sum^N_{i}\mathit{max}(0, \frac{AP(\bv{d}_{i})}{AP(\bv{d}^{tr}_{i})} - 1)
  \label{eq:rfloss}
\end{equation}
% \cref{fig: } illustrates the importance of $\mathcal{L}_{BOOST}$.
The final loss is the sum of the above two losses:
\begin{equation}
  \mathcal{L} = \mathcal{L}_{AP} + \lambda\mathcal{L}_{BOOST}
  \label{eq:loss}
\end{equation}
where $\lambda$ is a weight to regulate the second term. 
We use a differentiable approach (FastAP\cite{fastap}) to compute the Average Precision (AP) for each descriptor. 

Given a transformed descriptor $\bv{d}^{tr}_i\in\mathbb{R}^{1\times D}$ in the first image and the set of descriptors  $\mathbf{d'}^{tr}\in\mathbb{R}^{N\times D}$ in the second image. FastAP can be computed by using the ground truth labels about matched pairs $\mathbf{M}=\{M^+,M^-\}$ and pairwise distance vector $Z\in\mathbb{R}^{N}$ with value domain $\Omega$. 
By using distance quantization, $\Omega$ can be quantized as a finite set with $b$ elements $\Omega=\{z_1,z_2,\ldots,z_b\}$, then the precision and recall can be reformulated as functions of the distance $z$:
\begin{equation}
    \mathbf{Prec}(z) = P(M^+|Z<z)
\end{equation}
\begin{equation}
    \mathbf{Rec}(z) = P(Z<z|M^+)
\end{equation}
where $P(M^+|Z<z)$ represents the prior distribution for positive matches $M^+$ conditioned on $Z<z$ and $P(Z<z|M^+)$ is the cumulative distribution function (CDF) for $Z$.
Finally, the AP can be approximated by the area of precision-recall curve $\mathbf{PR}_z(\bv{d}^{tr}_i)=\{(\mathbf{Prec}(z), \mathbf{Rec}(z)), z\in\Omega\}$, which can be denoted as:
\begin{equation}
    \mathbf{FastAP} =  \int_{z\in \Omega}\mathbf{Prec}(z)d\mathbf{Rec}(z) 
    \label{eq:fastap}
\end{equation}
More details about FastAP are described in \cite{fastap}. The ground truth labels about matches $\mathbf{M}$ can be acquired using the ground truth poses and depth maps. Note that the way to calculate distance vector $Z$ is different for real-valued and binary descriptors.

\subsection{Different types of descriptors}
We are able to train our model to boost the descriptors into both binary and real-valued forms by using different ways to compute the distance vector $Z$.

\noindent
\textbf{Real-Valued Descriptors:}
We apply $L_2$ normalization to the output vector of the last layer of \M ,  and the pairwise distance vector $Z$ can be calculated as:
\begin{equation}
  Z = 2 - 2\bv{d}^{tr}_i(\mathbf{d'}^{tr})^\top
  \label{eq:euclidean}
\end{equation}
In this case, the bound range of $Z$ is $[0, 4]$ and we quantize the $\Omega$ as a finite set with 10 elements.

\noindent
\textbf{Binary Descriptors:}
We first use $tanh$ to threshold the output vector of the last layer of \M~to $[-1, 1]$. 
The output vector is then binarized to $\{-1, 1\}$.
However, there is no real gradient defined for binarization.
Our solution is to copy gradients from binarized vector to unbinarized vector following the straight-through estimator\cite{st}.
Finally, the pairwise distance vector $Z$ can be obtained as:
\begin{equation}
  Z = \frac{1}{2}(D - \bv{d}^{tr}_i(\mathbf{d'}^{tr})^\top)
  \label{eq:hamming}
\end{equation}
For the Hamming distance, the values of $Z$ are the integer in $\{0,1,\ldots,D\}$, and AP can be computed in a closed form by setting $b=D$ in FastAP. 
However, we use $b=10$ to get a larger margin between matching descriptors and non-matching descriptors as the discussion in \cite{fastap}.

\begin{figure*}[!t]
    \begin{minipage}{.6\linewidth}
    \centering
    \includegraphics[width=1.0\linewidth]{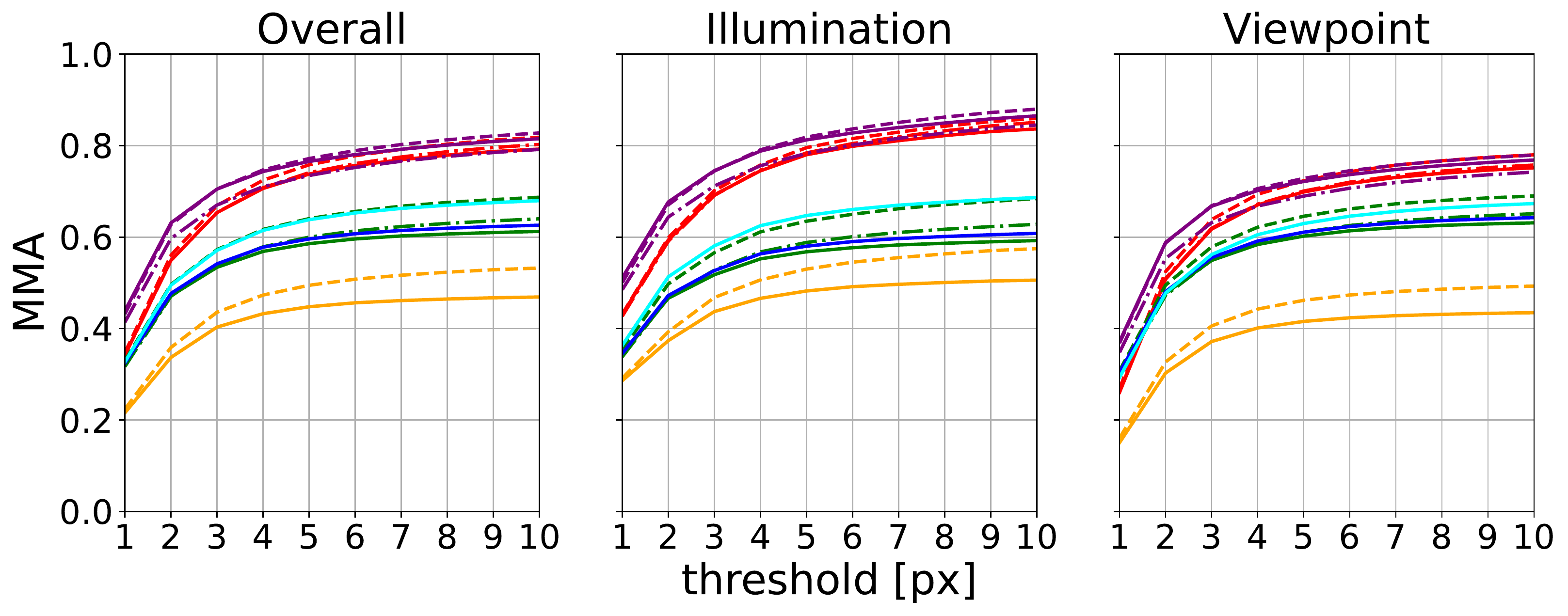}
    \end{minipage}
    \begin{minipage}{.4\linewidth}
    \centering
    \resizebox{1.0\columnwidth}{!}{
    % \resizebox{\textwidth}{15mm}{
    \scriptsize
    \begin{tabular}{lccc}
    \toprule
    \textbf{Method} & \textbf{Features} & \textbf{Matches} & \textbf{MMA @3 / @5} \\ \midrule
    \includegraphics[width=0.08\linewidth]{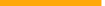} ORB\cite{orb}                         & 2956 & 997  & 0.403 / 0.448 \\
    \includegraphics[width=0.08\linewidth]{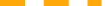} ORB+Boost-B (ours)          & 2956 & 1107 & \textbf{0.436} / \textbf{0.495} \\ \midrule
    \includegraphics[width=0.08\linewidth]{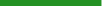} SIFT\cite{sift}                      & 1675 & 772  & 0.534 / 0.586 \\
    \includegraphics[width=0.08\linewidth]{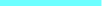} SOSNet\cite{sosnet}                & 1675 & 797  & \underline{0.571} / \underline{0.638} \\
    \includegraphics[width=0.08\linewidth]{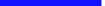} RootSIFT\cite{rootsift}          & 1675 & 799  & 0.542 / 0.596 \\
    \includegraphics[width=0.08\linewidth]{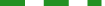} SIFT+Boost-F (ours)         & 1675 & 853  & \textbf{0.573} / \textbf{0.640} \\
    \includegraphics[width=0.08\linewidth]{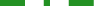} SIFT+Boost-B (ours)        & 1675 & 860  & 0.539 / 0.600 \\ \midrule
    \includegraphics[width=0.08\linewidth]{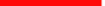} SuperPoint\cite{sp}                    & 1562 & 884  & \underline{0.654} / 0.738 \\
    \includegraphics[width=0.08\linewidth]{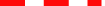} SuperPoint+Boost-F (ours)     & 1562 & 920  & \textbf{0.669} / \textbf{0.758} \\
    \includegraphics[width=0.08\linewidth]{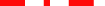} SuperPoint+Boost-B (ours)    & 1562 & 911  & \underline{0.654} / \underline{0.741} \\ \midrule
    \includegraphics[width=0.08\linewidth]{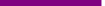} ALIKE\cite{alike}                   & 2578 & 1229 & \underline{0.705} / \underline{0.766} \\
    \includegraphics[width=0.08\linewidth]{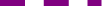} ALIKE+Boost-F (ours)       & 2578 & 1325 & \textbf{0.705} / \textbf{0.772} \\
    \includegraphics[width=0.08\linewidth]{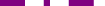} ALIKE+Boost-B (ours)       & 2578 & 1271 & 0.670 / 0.735 \\
    \bottomrule
    \end{tabular}}
    \end{minipage}
    \caption{MMA curves in HPatches (the higher the better) and the number of matched points on average (the larger the better). The results show that our feature booster can improve the performance for all the features. Boost-F and Boost-B indicate real-valued boosted and binary boosted descriptors, respectively.}
    \label{fig:hseq}
\end{figure*}

%-------------------Implementation details-------------------
\section{Implementation details}
\label{sec:implem}
In this section, we provide some implementation details for training \M .
\M~is plug-and-play and can be combined with any feature extraction process. In this paper, we trained \M s for ORB\cite{orb}, SIFT\cite{sift}, SuperPoint\cite{sp}, and ALIKE\cite{alike} respectively. We use ORB-SLAM2's\cite{orbslam2} extractor for ORB extraction and COLMAP's\cite{colmap1,colmap2} extractor for SIFT extraction. For SuperPoint\cite{sp}, we use its open-source repository and the Non-Maximum Suppression (NMS) radius is 4 pixels. For ALIKE\cite{alike}, we use its default open-source  model.

\noindent
\textbf{Architecture details:}
All the models were implemented in PyTorch\cite{pytorch}.
The Transformer in \M~uses $L = 9$ encoder layers for ALIKE and SuperPoint, and $L = 4$ for ORB and SIFT.
The query, key, and value in the Transformer encoder have the same dimension $D$ as that of the input descriptor.
The feed-forward network in Transformer is an MLP with 2 layers where the output dimensions are $(2D, D)$.
The geometric encoder is an MLP with five layers where the output dimensions are $(32, 64, 128, D, D)$ respectively.
Note the 2D locations of keypoints are normalized by the largest image dimension and the feature orientation is represented in radians.
For ORB (or binary) descriptors, we first convert them to a float vector and normalized them from $[0,1]$ to $[-1,1]$ and then send them to the 2-layer MLP with shortcut connection where the output dimensions are $(2D,D)$ like all other descriptors.

\noindent
\textbf{Training data:}
We trained all the \M s on MegaDepth\cite{megadepth} and adopt the training scenes used in DISK\cite{disk}. 
We computed the overlap score between two images following D2-Net\cite{d2net} and sampled 300 training pairs with an overlap score in $[0.1, 1]$ for each scene at every epoch.
A random $512 \times 512$ patch centered around one correspondence is selected for each pair.
During the training, all the local features were extracted on-the-fly, yielding up to 2048 local features from a single image.
The labels for matched descriptors and unmatched descriptors were generated by checking the distance between the re-projected points and the keypoints. For matched descriptors, the distance is below 3 pixels. For unmatched descriptors, the distance is greater than 15 pixels, considering the possible annotation errors.

\noindent
\textbf{Training details:}
We set $\lambda = 10$ in the training loss and trained our \M s using AdamW\cite{adamw} optimizer. We increased the learning rate to $1\times10^{-3}$ linearly in the first 500 steps and then decreased the learning rate in the form of cosine at each epoch in the following steps. The batch size is 16 during the training.

%-------------------Experiments-----------------------------
\section{Experiments}
\label{sec:exper}
\begin{figure*}[!t]
  \centering
%   \fbox{\rule{0pt}{2in} \rule{0.9\linewidth}{0pt}}
  \includegraphics[width=.9\linewidth]{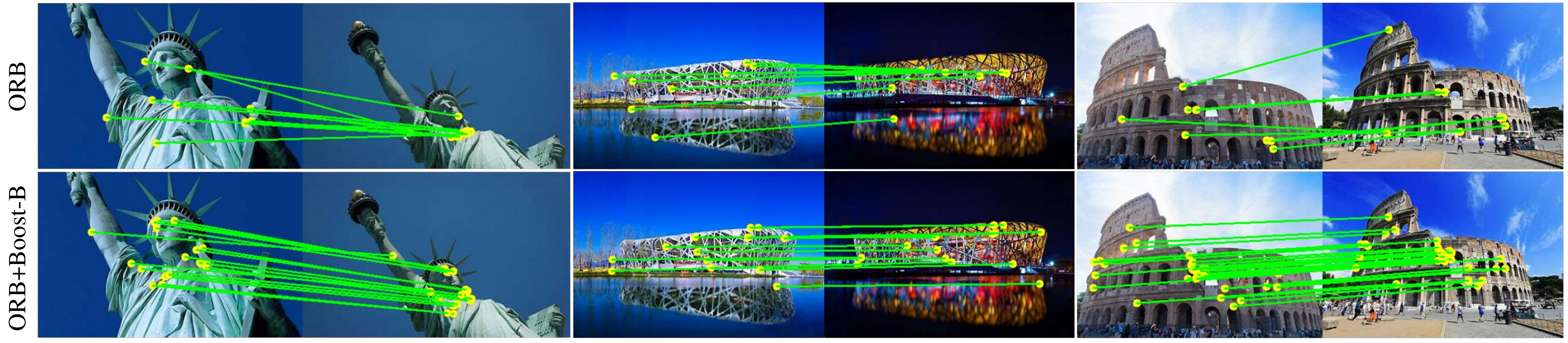}
  \caption{Matching results of using the original ORB\cite{orb} descriptors (\textbf{Top row}) and the boosted ORB descriptors (\textbf{Bottom row}) for Internet images. Nearest neighbor search and RANSAC\cite{ransac} were applied for matching.}
  \label{fig:realworld}
\end{figure*}

After training our model on MegaDepth\cite{megadepth}, we evaluate the trained model on image matching, visual localization, and structure-from-motion tasks using the public benchmark datasets. Note we do not fine-tune the model using the images from those datasets. We also show some matching results for real-world images from the Internet in \cref{fig:realworld}. Finally, we also conduct an ablation study about the key components of our method.

\begin{table*}[ht]
\begin{minipage}{.6\linewidth}
\centering
\resizebox{1.\columnwidth}{!}{
\scriptsize
\begin{tabular}{lcccc}
\toprule
    & \multicolumn{2}{c}{\textbf{Aachen Day-Night V1.1}\cite{aachenv1.1}} & \multicolumn{2}{c}{\textbf{InLoc}\cite{inloc}} \\ 
    & \multicolumn{2}{c}{\underline{(0.25m,2$^{\circ}$) / (0.50m,5$^{\circ}$) / (5.0m,10$^{\circ}$) $\uparrow$}} &
      \multicolumn{2}{c}{\underline{(0.25m,10$^{\circ}$) / (0.50m,10$^{\circ}$) / (5.0m,10$^{\circ}$) $\uparrow$}} \\ 
\multirow{-3}{*}{\textbf{Method}} &  Day   & Night   & DUC1   & DUC2                       \\
\midrule
ORB\cite{orb}                        & 80.6 / 87.9 / 93.6 & 31.9 / 37.2 / 49.2  & 24.7 / 33.3 / 42.4 & 26.7 / 37.4 / 44.3 \\
ORB-Boost-B (Ours)        & \textbf{83.1} / \textbf{89.8} / \textbf{94.7} & \textbf{49.2} / \textbf{61.8} / \textbf{73.3}  & \textbf{35.4} / \textbf{50.5} / \textbf{59.1} & \textbf{38.9} / \textbf{51.9} / \textbf{61.8}  \\
\midrule
SIFT\cite{sift}             & 87.1 / 93.8 / 98.1 & 50.8 / 70.2 / 81.2  & 29.3 / 43.4 / 51.5 & 	19.1 / 33.6 / 40.5  \\
SOSNet\cite{sosnet}    & \textbf{88.7} / \textbf{94.7} / \textbf{98.7} & 58.1 / \textbf{78.5} / \textbf{92.7}  & \textbf{35.9} / \textbf{50.0} / \textbf{64.6} & \underline{26.7} / \textbf{43.5} / \textbf{56.5}  \\
RootSIFT\cite{rootsift}     & 86.8 / 94.1 / \underline{98.4} & 57.1 / 76.4 / 88.5  & 30.3 / 46.5 / 57.1 & 22.1 / \underline{42.7} / 50.4  \\
SIFT+Boost-F (Ours)        & 87.1 / \underline{94.5} / 98.1 & \underline{62.3} / \underline{78.0} / \underline{92.1}  & 31.8 / 43.9 / 57.1 & 24.4 / 36.6 / 49.6  \\
SIFT+Boost-B (Ours)       & \underline{87.5} / \underline{94.5} / 98.1 & \textbf{63.9} / 77.5 / 91.1  & \underline{32.8} / \underline{47.5} / \underline{57.6} & \textbf{30.5} / \textbf{43.5} / \underline{51.1}  \\
\midrule
SuperPoint\cite{sp}         & \underline{87.9} / \underline{94.3} / \underline{98.2} & 67.0 / \underline{84.8} / 95.8  & \underline{36.9} / \underline{57.6} / 64.6 & \underline{38.2} / 55.0 / 65.6  \\
SuperPoint+Boost-F (Ours)  & \textbf{88.3} / \textbf{94.4} / \textbf{98.7} & \textbf{70.2} / \textbf{85.9} / \textbf{97.9}  & \textbf{41.4} / \textbf{58.6} / \textbf{69.2} & \textbf{40.5} / \textbf{58.0} / \textbf{67.9}  \\
SuperPoint+Boost-B (Ours) & 87.4 / 94.1 / 97.9 & \underline{68.6} / \underline{84.8} / \underline{96.3}  & \underline{36.9} / 54.5 / \underline{65.7} & 35.9 / \textbf{58.0} / \textbf{67.9}  \\
\midrule
ALIKE\cite{alike}          & \textbf{87.3} / 93.2 / \underline{98.7} & 67.5 / 85.3 / \underline{97.9}  &  29.3 / 46.5 / 59.6  & 25.2 / 38.9 / 47.3 \\
ALIKE+Boost-F (Ours)       & 86.7 / \textbf{94.2} / \textbf{99.0} & \textbf{72.8} / \textbf{86.9} / \textbf{98.4}  & \underline{35.4} / \underline{51.0} / \underline{65.7}  & \underline{29.8} / \underline{44.3} / \underline{55.7}  \\
ALIKE+Boost-B (Ours)       & \underline{86.9} / \underline{93.8} / 98.3 & \underline{71.7} / \underline{86.4} / 96.9  & \textbf{35.9} / \textbf{54.0} / \textbf{66.2}  & \textbf{30.5} / \textbf{49.6} / \textbf{63.4}  \\
\midrule
SuperPoint+SuperGlue\cite{sp,sg} & 89.6 / 96.4 / 99.3 & 73.3 / 90.6 / 100.0 &  44.9 / 64.6 / 78.3 & 49.6 / 73.3 / 77.1 \\ 
\bottomrule
\end{tabular}
}
\end{minipage}
\begin{minipage}{.4\linewidth}
  \centering
%   \fbox{\rule{0pt}{2in} \rule{0.9\linewidth}{0pt}}
  \includegraphics[width=.95\linewidth]{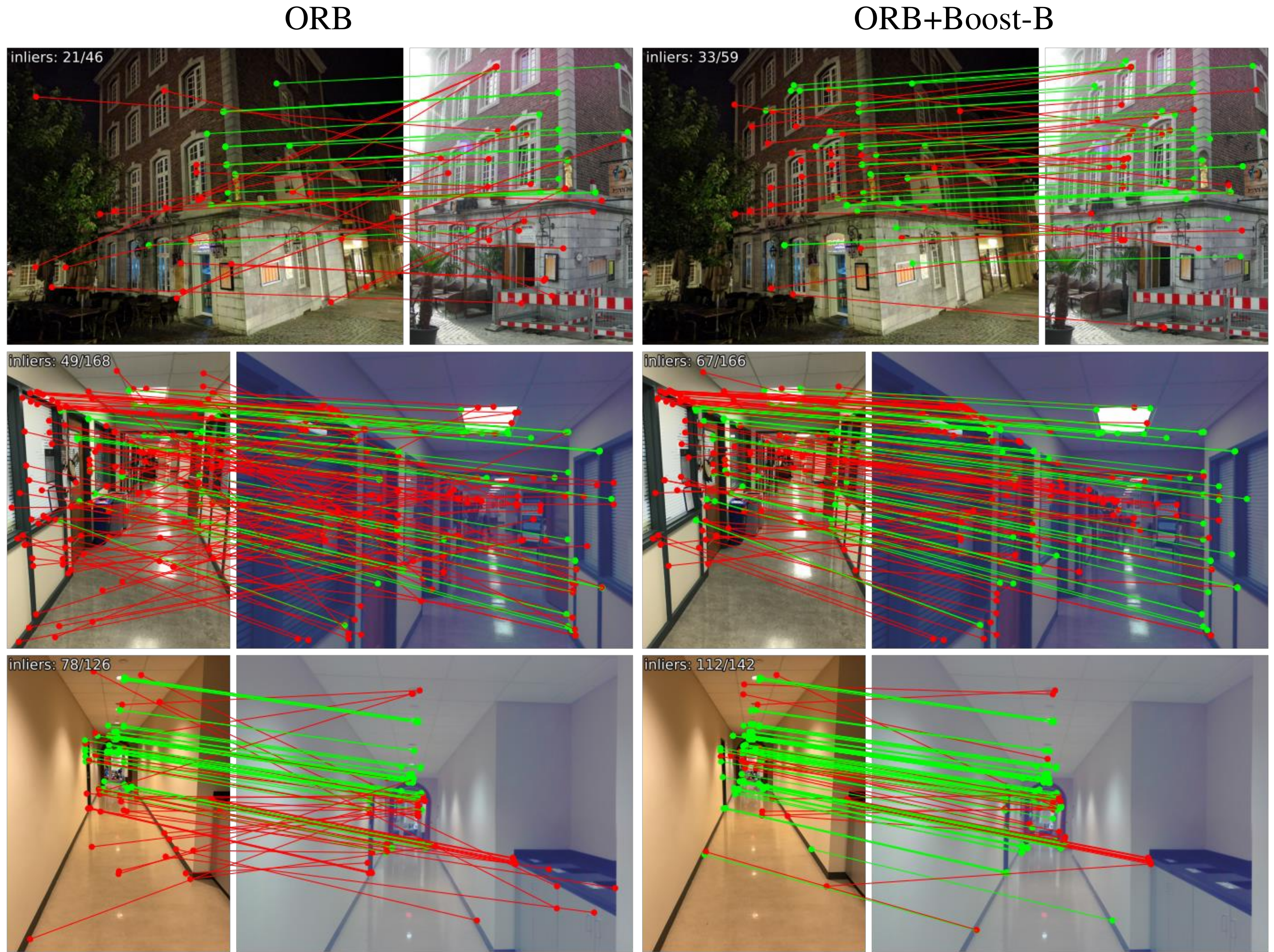}
%   \caption{Visualization of matching.}
  \label{fig:visloc-matches}
\end{minipage}
\caption{Visual localization results in both outdoor (Aachen Day-Neight\cite{aachenv1.1}) and indoor scenes (InLoc\cite{inloc}). The positional and angular performances are present (the larger the better). Note that the boosted ORB (ORB-Boost-B) even outperforms ALIKE\cite{alike} and can compete with SuperPoint\cite{sp} in indoor scenes.
Images on the right show some matching results before and after boosting using ORB\cite{orb} descriptors (red lines indicate wrong correspondences). }
\label{tab:visloc}
\end{table*}

% \begin{figure}[!t]
%   \centering
% %   \fbox{\rule{0pt}{2in} \rule{0.9\linewidth}{0pt}}
%   \includegraphics[width=1.0\linewidth]{figure/visloc-matches.pdf}
%   \caption{Visualization of matching.}
%   \label{fig:visloc-matches}
% \end{figure}

%--------------------------------
\subsection{Image Matching}
We first evaluate our method on the image matching task using the HPatches\cite{hpatches} test sequences. 
HPatches dataset contains 116 different sequences of which 58 sequences have illumination changes and 58 sequences have viewpoint changes. 
Following D2Net\cite{d2net}, we excluded eight sequences for this experiment.

\noindent
\textbf{Experiment setup:}
We follow the evaluation protocol in D2Net\cite{d2net} and record the mean matching accuracy
(MMA)\cite{MMA} under thresholds varying from 1 to 10 pixels, together with the numbers of features and matches.
The MMA is defined as the average percentage of correct matches under different reprojection error thresholds.
Like D2-Net, we use mutual nearest neighbor search as the matching method.
For comparison, we report the results of raw descriptors, boosted descriptors by our approach, a variant for SIFT (RootSIFT\cite{rootsift}), and a learning-based patch descriptor (SOSNet\cite{sosnet}). All the DoG-based descriptors were computed from the same DoG keypoints for a fair comparison.

\noindent
\textbf{Result:}
\cref{fig:hseq} shows MMA results on HPatches under illumination and viewpoint change.
Our method can enhance the performance of all descriptors for either the transformed real-valued descriptors or the binary ones. 
For SIFT, the transformed real-valued descriptors by our method outperforms SOSNet, while can find more correct matches as shown in the Table as shown in \cref{fig:hseq}. In addition, we can see the potential of \M~for descriptor compression (real-valued descriptor to binary descriptor). The transformed binary descriptor from SuperPoint has a similar performance to the original SuperPoint under both illumination and viewpoint change while producing more correct matches. It is also interesting to see that the binary descriptor boosted from SIFT performs better than both SIFT and RootSIFT.

%--------------------------------
\subsection{Visual Localization}
In the second experiment, we evaluate our method in visual localization, a more complete pipeline in computer vision. Two challenging scenarios are selected for evaluation: an outdoor dataset with severe illumination changes and a large-scale indoor dataset with plenty of texture-less areas and repetitive patterns.

\noindent
\textbf{Experiment setup:}
For the outdoor scenes, we use the Aachen Day-Night dataset v1.1\cite{aachenv1.1}, which contains 6697 day-time database images and 1015 query images (824 for the day and 191 for the night). 
For the indoor scenes, we use the InLoc dataset\cite{inloc}, which contains about 10k database images collected in two buildings. 
We use the hierarchical localization toolbox (HLoc)\cite{hloc} for visual localization on Aachen Day-Night and InLoc dataset by replacing the feature extraction module with different feature detectors and descriptors. 
We use the evaluation protocol on the Long-Term Visual Localization Benchmark\cite{visloc} and report the percentage of correct localized query images under given error thresholds. For comparison, we also report the result of the learning-based matching method (SuperPoint+SuperGlue). Not that all other methods use mutual nearest neighbor search for matching. We adopt ratio test or distance test for mutual nearest neighbor matching.  For a fair comparison, the ratio or distance thresholds of all the transformed descriptors are selected according to the threshold criteria of their corresponding baselines\footnote{Please see the supplementary material for additional details.}.

\noindent
\textbf{Result:}
The results are shown in \cref{tab:visloc}. Our method significantly improves the performance for all the features in both outdoor and indoor environments, especially for SIFT. After boosting, even the binary ORB descriptors can compete with the SuperPoint and outperform ALIKE in indoor environments (InLoc).
We can see that the real-valued and binary boosted SIFT both show considerable competitiveness compared to SOSNet on the Day-Night outdoor dataset.
The result also can show that SuperGlue still has the best performance in this experiment. However, our method boosts descriptors before the matching stage, making it more versatile and easy to insert into existing systems.

%--------------------------------
\subsection{Structure-from-motion}
\noindent
\textbf{Experiment setup:}
We use three medium-scale datasets in the ETH SfM benchmark\cite{sfm} following D2-Net\cite{d2net} for evaluation.
We use exhaustive image matching for all these datasets and adopt ratio test or distance test for mutual nearest neighbor matching. 
Then, we run the SfM using COLMAP\cite{colmap1,colmap2}.
Following the evaluation protocol defined by \cite{sfm}, we report the number of registered images, sparse points, total observations in image, mean feature track length, and mean re-projection error.

\noindent
\textbf{Result:}
\cref{tab:sfm} shows the results. Our approach again enhances the performance of all the features on the task of structure-from-motion. 
Our method can help the original features to produce a more complete reconstruction, as our approach can register more images and reconstruct more 3D points as shown in \cref{tab:sfm}.
Besides, our \M~can achieve higher feature track length, which means that we can find more correspondences between images to reconstruct 3D points while tracking the same features across more images.
We also observe the situation that has been discussed in \cite{geodesc,sosnet} that more matches tend to lend higher re-projection error, and we think this issue can be addressed by recent work on keypoint position refinement\cite{patchflow,pixsfm}.

\begin{table}
\centering
\resizebox{1.0\columnwidth}{!}{
\Huge
\begin{tabular}{llccccc}
\toprule
\textbf{Dataset} &
  \textbf{Descriptor} &
  \begin{tabular}[c]{@{}c@{}}{\textbf{\#Reg.}}\\ {\textbf{Images} $\uparrow$}\end{tabular} &
  \begin{tabular}[c]{@{}c@{}}{\textbf{\#Sparse.}}\\ {\textbf{Point} $\uparrow$}\end{tabular} &
  {\textbf{\#Obs.} $\uparrow$} &
  \begin{tabular}[c]{@{}c@{}}{\textbf{\#Track}}\\ {\textbf{Length} $\uparrow$}\end{tabular} &
  \begin{tabular}[c]{@{}c@{}}{\textbf{\#Reproj.}}\\ {\textbf{Error} $\downarrow$}\end{tabular} \\
\midrule
\multirow{10}{*}{\begin{tabular}[c]{@{}l@{}}{\textbf{Madrid}}\\ {\textbf{Metropolis}}\\ 1344 images\end{tabular}} 
                                                            %                                 & ORB                  
                                                            %  &      &     &      &      &        \\
                                                            %                                 & ORB-Boost-B        &      &     &      &      &        \\
                                                                                            & SIFT\cite{sift}             & 417  & 29653 & 210460 & 7.10 & \textbf{0.78px} \\
                                                                                            & SOSNet\cite{sosnet}    & \textbf{464}  & \textbf{35288} & \textbf{260737} & \textbf{7.39} & 0.87px \\
                                                                                            & RootSIFT\cite{rootsift}     & \underline{443}  & 32613 & 230487 & 7.07 & \underline{0.79px} \\
                                                                                            & SIFT+Boost-B (ours)       & 415  & \underline{34497} & \underline{242053} & 7.02 & 0.86px \\
                                                                                            & SIFT+Boost-F (ours)        & 409  & 30020 & 221320 & \underline{7.37} & 0.88px \\
                                                                                     \cline{2-7}
                                                                                            & SuperPoint\cite{sp}         & \underline{512}  & \underline{29131} & \underline{230966} & 7.93 & \textbf{1.14px} \\
                                                                                            & SuperPoint+Boost-B (ours) & 433  & 25872 & 218370 & \textbf{8.44} & \underline{1.18px} \\
                                                                                            & SuperPoint+Boost-F (ours)  & \textbf{534}  & \textbf{34033} & \textbf{276204} & \underline{8.12} & 1.19px \\
\midrule
\multirow{10}{*}{\begin{tabular}[c]{@{}l@{}}{\textbf{Gendarmen-}}\\ {\textbf{markt}}\\ 1463 images\end{tabular}}  
                                                            %                                 & ORB          
                                                            %  &      &     &      &      &        \\
                                                            %                                 & ORB-Boost-B        &      &     &      &      &        \\
                                                                                            & SIFT\cite{sift}             & 944  & 75369 & 476495 & 6.32 & \textbf{0.91px} \\
                                                                                            & SOSNet\cite{sosnet}    & \textbf{972}  & \underline{85507} & \textbf{591623} & \textbf{6.92} & 1.00px \\
                                                                                            & RootSIFT\cite{rootsift}     & \underline{955}  & 77888 & 511209 & \underline{6.56} & \underline{0.93px} \\
                                                                                            & SIFT+Boost-B (ours)       & 944  & \textbf{95537} & \underline{581878} & 6.09 & 0.99px \\
                                                                                            & SIFT+Boost-F (ours)        & 937  & 84496 & 552081 & 6.53 & 1.01px \\
                                                                                     \cline{2-7}
                                                                                            & SuperPoint\cite{sp}         & \underline{997}  & \underline{70971} & \underline{535761} & 7.55 & \textbf{1.18px} \\
                                                                                            & SuperPoint+Boost-B (ours) & 951  & 62426 & 513442 & \textbf{8.22} & 1.23px \\
                                                                                            & SuperPoint+Boost-F (ours)  & \textbf{1044} & \textbf{84052} & \textbf{635591} & \underline{7.56} & \underline{1.20px} \\
\midrule
\multirow{10}{*}{\begin{tabular}[c]{@{}l@{}}{\textbf{Tower of}}\\ {\textbf{London}}\\ 1576 images\end{tabular}}   
                                                            %                                 & ORB       
                                                            %  &      &     &      &      &        \\
                                                            %                                 & ORB-Boost-B        &      &     &      &      &        \\
                                                                                            & SIFT\cite{sift}             & 667  & 61906 & 457193 & 7.39 & \textbf{0.78px} \\
                                                                                            & SOSNet\cite{sosnet}    & \textbf{738}  & \underline{71734} & \textbf{558944} & \textbf{7.79} & 0.84px \\
                                                                                            & RootSIFT\cite{rootsift}     & 674  & 62348 & 472817 & \underline{7.58} & \underline{0.79px} \\
                                                                                            & SIFT+Boost-B (ours)       & \underline{690}  & \textbf{73954} & \underline{515206} & 6.97 & 0.82px \\
                                                                                            & SIFT+Boost-F (ours)        & 681  & 66309 & 491273 & 7.41 & 0.83px \\
                                                                                     \cline{2-7}
                                                                                            & SuperPoint\cite{sp}         & \underline{712}  & \underline{38921} & \underline{313825} & \underline{8.06} & \textbf{1.12px} \\
                                                                                            & SuperPoint+Boost-B (ours) & 653  & 34641 & 290505 & \textbf{8.39} & \underline{1.14px} \\
                                                                                            & SuperPoint+Boost-F (ours)  & \textbf{773}  & \textbf{45687} & \textbf{360642} & 7.89 & \underline{1.14px} \\ 
\bottomrule   
\end{tabular}
}
\caption{Results on structure-from-motion. Our method improves the performance of the existing descriptors (SIFT\cite{sift}, and SuperPoint\cite{sp}) in three datasets of ETH SfM benchmark\cite{sfm}.}
\label{tab:sfm}
\end{table}

% \begin{figure}[!t]
%   \centering
% %   \fbox{\rule{0pt}{2in} \rule{0.9\linewidth}{0pt}}
%   \includegraphics[width=1.0\linewidth]{figure/hseq-ablation.pdf}
%   \caption{HPatches ablation study.}
%   \label{fig:hseq-ablation}
% \end{figure}

% \begin{figure}[!t]
%     \begin{minipage}{.4\linewidth}
%     \centering
%     \includegraphics[width=0.95\linewidth]{figure/hseq-ablation.pdf}
%     \end{minipage}
%     \begin{minipage}{.55\linewidth}
%     \centering
%     \resizebox{1.0\columnwidth}{!}{
%     % \resizebox{\textwidth}{15mm}{
%     \begin{tabular}{llccc}
%     \toprule
%     \multicolumn{2}{l}{\textbf{Method}} & \textbf{\#Matches} & \textbf{\#MMA@3} & \textbf{\#MMA@5} \\
%     \midrule
%     \multirow{4}{*}{SuperPoint} & No Boost          & 975  & 0.648 & 0.731   \\
%                                 & Only DescEncoder  & 980  & 0.650 & 0.734 \\
%                                 & No Cross-boosting & 986  & 0.655 & 0.740   \\
%                                 & Full              & \textbf{1012} & \textbf{0.661} & \textbf{0.748} \\
%     \bottomrule
%     \end{tabular}
%     }
%     \end{minipage}
%     \caption{Ablation study of \M~for SuperPoint in HPatches (the higher the better). The results show that cross-boosting can significantly improve the performance.}
%     \label{fig:hseq-ablation}
% \end{figure}

\begin{table}
\centering
\resizebox{1.0\columnwidth}{!}{
    % \begin{tabular}{llccc}
    % \toprule
    % \multicolumn{2}{l}{\textbf{Descriptor}} & \textbf{\#Matches} & \textbf{\#MMA@3} & \textbf{\#MMA@5} \\
    % \midrule
    % \multirow{4}{*}{SuperPoint} & No Boost          & 975  & 0.648 & 0.731   \\
    %                             & Only DescEncoder  & 980  & 0.650 & 0.734 \\
    %                             & Only Self-boosting & 986  & 0.655 & 0.740   \\
    %                             & Full (Self+Cross-boosting)              & \textbf{1012} & \textbf{0.661} & \textbf{0.748} \\
    % \bottomrule
    % \end{tabular}
    \begin{tabular}{lccccc}
    \toprule
    \multirow{2}{*}{\textbf{Descriptor}} &
      \multicolumn{2}{c}{\textbf{Self Boosting}} &
      \multirow{2}{*}{\textbf{Cross Boosting}} &
    %   \multirow{2}{*}{\begin{tabular}[c]{@{}c@{}}\textbf{Cross}\\ \textbf{Boosting}\end{tabular}} &
      \multirow{2}{*}{\begin{tabular}[c]{@{}c@{}}\textbf{HPatches}\\ \textbf{Matches}\end{tabular}} &
      \multirow{2}{*}{\begin{tabular}[c]{@{}c@{}}\textbf{HPatches MMA}\\ \textbf{@3 / @5}\end{tabular}} \\ \cline{2-3}
                                & $\mathbf{MLP}_{desc}$ & $\mathbf{MLP}_{geo}$ &              &     &                                 \\
    \midrule
    \multirow{5}{*}{SuperPoint\cite{sp}} &                       &                      &              & 883 & 0.654 / 0.738                   \\
                                & $\checkmark$          &                      &              & 883 & 0.654 / 0.738                   \\
                                & $\checkmark$          & $\checkmark$         &              & 884 & 0.655 / 0.739                   \\
                                & $\checkmark$          &                      & $\checkmark$ & 893 & 0.657 / 0.742                   \\
                                & $\checkmark$          & $\checkmark$         & $\checkmark$ & 919 & \textbf{0.669} / \textbf{0.758} \\
    \bottomrule
    \end{tabular}
}
    \caption{Ablation study on SuperPoint\cite{sp} in HPatches\cite{hpatches} (the higher the better). The results show that cross-boosting can significantly improve the performance.}
    \label{tab:hseq-ablation}
\end{table}

%--------------------------------
\subsection{Ablation Study}
\cref{tab:hseq-ablation} shows an ablation study of different components in our network. The study shows that geometric encoding is necessary for self-boosting, and the cross-boosting has a better performance for descriptor boosting. With the help of both modules, our transformed descriptors perform significantly better.

%-------------------Discussion-----------------------------
\section{Discussion}
\label{sec:discus}
\noindent
\textbf{Computational cost:} Our network is lightweight and efficient.
We measure the runtime of our method on both a desktop GPU and an embedded GPU. 
A forward pass with 2000 features in NVIDIA RTX 3090 takes on average 3.2/4.7ms for our 4/9 layers network, while in NVIDIA Jetson Xavier NX it needs 27/46ms.

\noindent
\textbf{Generalization:} Though for each feature we need to train their corresponding \M, experiments show that our approach works well for various classes of descriptors (hand-crafted or learned, binary or real-valued). Our models are trained with the MegaDepth\cite{megadepth} dataset and do not need to be fine-tuned for different tasks or datasets. 

\noindent
\textbf{Limitations:} The performance of the boosted descriptor is limited by the representation ability of the raw descriptor, though the performance gain tends to be larger for weaker descriptors like ORB. Our approach cannot be applied to enhance dense features because the computational cost grows with the number of feature points.

%-------------------Conclusion-----------------------------
\section{Conclusion}
\label{sec:conclude}
We introduce a descriptor enhancement stage into the traditional feature matching pipeline and propose a versatile and lightweight framework for descriptor enhancement called \M. \M~jointly processes the geometric properties and visual descriptors of all the keypoints within a single image to extract the global contextual information. With the help of the global context, the transformed descriptors become powerful even though the original descriptor is very weak. Our experiments show that \M~can help various classes of descriptors (SIFT, ORB, SuperPoint, and ALIKE) to perform better under different vision tasks. Furthermore, our \M~demonstrates its potential for descriptor compression and can run in real time. We believe that our \M~can be useful for many practical applications.

\ifarxiv \section*{\Large Supplementary Material}
\appendix
\label{sec:appendix}

This supplementary material provides the following additional information: 
Section \ref{sec:indoor} presents the result of indoor visual localization using NN search with the mutual check. 
Section \ref{sec:slam} provides the result of our method in visual SLAM.  
Section \ref{sec:efficiency} shows the efficiency of different Transformer modules for cross-boosting stage.
Section \ref{sec:loss-abla} provides an ablation study of the loss function used to train our method. 
As mentioned in Section 5.2 in the paper, Section \ref{sec:threshold} details how we chose the threshold for Lowe’s ratio test\cite{sift} or distance test used for the visual localization and 3D reconstructions. 
Section \ref{sec:example} shows more qualitative examples of the matching results of our approach (before and after boosting) on the Aachen Day-Night v1.1\cite{aachenv1.1} and InLoc\cite{inloc} datasets.

\section{Indoor visual localization}
\label{sec:indoor}

\begin{table}[h]
    \centering
    \resizebox{1.\columnwidth}{!}{
    \scriptsize
    % \Huge
    % \begin{tabular}{lcc}
    % \toprule
    %     & \multicolumn{2}{c}{\textbf{InLoc}} \\ 
    %     & \multicolumn{2}{c}{(0.25m,10$^{\circ}$) / (0.50m,10$^{\circ}$) / (5.0m,10$^{\circ}$) $\uparrow$} \\ \cline{2-3} 
    % \multirow{-3}{*}{\textbf{Method}} &  DUC1   & DUC2      \\
    % \midrule
    % ORB\cite{orb}             & \textbf{23.2} / 27.8 / 35.4 & 18.3 / 25.2 / 31.3 \\
    % ORB+Boost-B (Ours)        & 22.2 / \textbf{32.3} / \textbf{39.9} & \textbf{22.9} / \textbf{31.3} / \textbf{37.4} \\
    % \midrule
    % SIFT\cite{sift}           & 24.7 / 32.8 / 42.4 & 16.8 / 25.2 / 29.0 \\
    % SOSNet\cite{sosnet}       & \textbf{32.8} / \textbf{44.4} / \textbf{54.5} & \textbf{22.1} / \textbf{32.1} / \textbf{40.5} \\
    % RootSIFT\cite{rootsift}   & 24.2 / 33.8 / 41.4 & 16.8 / 26.0 / 32.1 \\
    % SIFT+Boost-F (Ours)       & \underline{26.8} / \underline{38.4} / \underline{47.5} & 17.6 / 29.8 / 34.4 \\
    % SIFT+Boost-B (Ours)       & 24.7 / 35.4 / 43.9 & \underline{19.1} / \underline{31.3} / \underline{35.9} \\
    % \midrule
    % SuperPoint\cite{sp}       & 33.3 / 49.5 / 61.1 & \underline{33.6} / \underline{51.9} / \underline{61.8} \\
    % SuperPoint+Boost-F (Ours) & \underline{33.8} / \textbf{54.5} / \textbf{65.2} & \textbf{35.9} / \textbf{57.3} / \textbf{63.4} \\
    % SuperPoint+Boost-B (Ours) & \textbf{35.9} / \underline{51.5} / \underline{61.6} & 29.0 / 48.1 / 58.0 \\
    % \midrule
    % ALIKE\cite{alike}         & \textbf{33.8} / 49.0 / 59.6 & 27.5 / 42.7 / 48.9 \\
    % ALIKE+Boost-F (Ours)      & \textbf{33.8} / \textbf{49.5} / \textbf{62.6} & \textbf{32.1} / \textbf{45.8} / \textbf{53.4} \\
    % \bottomrule
    % \end{tabular}
    \begin{tabular}{lcc}
    \toprule
        & \multicolumn{2}{c}{\textbf{InLoc}\cite{inloc}} \\ 
        & \multicolumn{2}{c}{\underline{(0.25m,10$^{\circ}$) / (0.50m,10$^{\circ}$) / (5.0m,10$^{\circ}$) $\uparrow$}} \\
    \multirow{-3}{*}{\textbf{Method}} &  DUC1   & DUC2      \\
    \midrule
    ORB\cite{orb}             & 21.7 / 30.8 / 36.9 & \textbf{24.4} / \textbf{30.5} / 35.9 \\
    ORB+Boost-B (Ours)        & \textbf{25.3} / \textbf{36.4} / \textbf{43.4} & 23.7 / 29.8 / \textbf{37.4} \\
    \midrule
    SIFT\cite{sift}           & 23.2 / 35.9 / 46.0 & 13.0 / 22.1 / 28.2 \\
    SOSNet\cite{sosnet}       & \textbf{31.8} / \textbf{44.4} / \textbf{54.0} & \textbf{23.7} / \textbf{39.7} / \textbf{48.1} \\
    RootSIFT\cite{rootsift}   & 24.7 / 36.9 / 41.9 & 17.6 / 27.5 / 33.6 \\
    SIFT+Boost-F (Ours)       & \underline{28.3} / \underline{40.4} / \underline{47.5} & \underline{19.8} / \underline{29.0} / \underline{35.1} \\
    SIFT+Boost-B (Ours)       & 24.2 / 35.9 / 46.0 & 18.3 / \underline{29.0} / \underline{35.1} \\
    \midrule
    SuperPoint\cite{sp}       & \textbf{33.3} / \underline{49.5} / \underline{61.1} & 33.6 / 51.9 / \textbf{61.8} \\
    SuperPoint+Boost-F (Ours) & \underline{32.3} / \textbf{51.0} / \textbf{64.1} & \textbf{36.6} / 51.9 / \underline{59.5} \\
    SuperPoint+Boost-B (Ours) & \textbf{33.3} / 49.0 / 60.1 & \underline{35.1} / 51.9 / \underline{59.5} \\
    \midrule
    ALIKE\cite{alike}         & \underline{31.8} / \underline{47.5} / \underline{61.1} & \underline{26.7} / \underline{41.2} / \textbf{49.6} \\
    ALIKE+Boost-F (Ours)      & \textbf{33.8} / \textbf{53.0} / \textbf{68.2} & \textbf{31.3} / \textbf{42.0} / \underline{48.1} \\
    ALIKE+Boost-B (Ours)      & 28.8 / 43.9 / 56.6 & \textbf{31.3} / 39.7 / 45.8 \\
    \bottomrule
    \end{tabular}
    }
    \caption{Visual localization results in indoor scenes (InLoc\cite{inloc}). The \textbf{first} and \underline{second} best result are highlighted. In this test, no ratio/distance test is used for feature matching.}
    \label{tab:inloc}
\end{table}

To further evaluate the performance of our method, we apply our method for visual localization on the InLoc dataset\cite{inloc} using only NN search and a mutual check without using the ratio or distance tests. 

As shown in \cref{tab:inloc}, our method can also enhance the performance of all descriptors although a different matching strategy is used. 
The SIFT+Boost-B is better than both SIFT\cite{sift} and RootSIFT\cite{rootsift}. 
The SuperPoint+Boost-B shows considerable competitiveness compared with SuperPoint\cite{sp}. 
We can also see that our ORB+Boost-B performs worse compared with SuperPoint\cite{sp} and ALIKE\cite{alike} without distance tests. 
In comparison, the results in Section 5.2 in the paper show that our ORB+Boost-B can compete with SuperPoint and ALIKE when we adopt ratio or distance tests for matching.

\section{Visual SLAM}
\label{sec:slam}
Our approach of reusing existing descriptors offers a cost-effective way to enhance the performance of established systems like visual SLAM. To demonstrate this, we integrated our ORB+Boost-B into ORB-SLAM2 \cite{orbslam2}.

The results of translation error in EuRoC dataset\cite{euroc} for ORB-SLAM2\cite{orbslam2} using ORB and ORB+Boost-B are shown in \cref{tab:slam}. By boosting the original ORB \cite{orb} to ORB+Boost-B, ORB-SLAM2 provides more accurate estimate. Compared to other state-of-the-art local features, our method can improve the performance while introducing minimal additional time consumption (only 3.2ms on a desktop GPU and 27ms on an embedded GPU to process 2000 ORBs).

\begin{table*}[]
    \centering
    \resizebox{1.\linewidth}{!}{
    \scriptsize
    \begin{tabular}{lccccccccccc}
    \toprule
    \textbf{Descriptor used} (ORB-SLAM2\cite{orbslam2}) & MH01 & MH02 & MH03 & MH04 & MH05 & V101 & V102 & V103 & V201 & V202 & V203 \\
    \midrule
    ORB\cite{orb} &
      0.0318 &
      0.0215 &
      0.0267 &
      0.1282 &
      0.0549 &
      0.0349 &
      \textbf{0.0211} &
      0.0486 &
      0.0449 &
      0.0270 &
      \textbf{0.1716} \\
    ORB+Boost-B (Ours) &
      \textbf{0.0304} &
      \textbf{0.0175} &
      \textbf{0.0252} &
      \textbf{0.0916} &
      \textbf{0.0470} &
      \textbf{0.0343} &
      0.0213 &
      \textbf{0.0449} &
      \textbf{0.0379} &
      \textbf{0.0249} &
      0.2606         \\
    \bottomrule
    \end{tabular}
    }
    \caption{Comparison of translation RMSE(m) in EuRoC dataset\cite{euroc} for ORB-SLAM2\cite{orbslam2} using different descriptors. RMSE is the smaller the better and the \textbf{better} results are highlighted. The result shows that ORB+Boost-B improves the accuracy of ORB-SLAM2.}
    \label{tab:slam}
\end{table*}

\section{Transformer modules for cross-boosting}
\label{sec:efficiency}

\begin{table*}[]
\centering
    \resizebox{1.0\linewidth}{!}{
    \scriptsize
    \begin{tabular}{llccc}
    \toprule
    % \hline
     & & & \\
    \multirow{-2}{*}{\textbf{Descriptor}} &
    \multirow{-2}{*}{\textbf{Module used} (Cross-boosting)} & \multirow{-2}{*}{\begin{tabular}[c]{@{}c@{}}{\textbf{HPatches MMA}$\uparrow$}\\ @3 / @5\end{tabular}} & \multirow{-2}{*}{\begin{tabular}[c]{@{}c@{}}{\textbf{RTX 3090 Runtime}(ms)$\downarrow$}\\ \#500 / \#1000 / \#2000 / \#4000 / \#8000\end{tabular}} & 
    \multirow{-2}{*}{\begin{tabular}[c]{@{}c@{}}{\textbf{Jetson NX Runtime}(ms)$\downarrow$}\\ \#500 / \#1000 / \#2000 / \#4000 / \#8000\end{tabular}} \\
    \midrule
    % \hline
    % ORB+Boost-B
                                  & Vanilla Transformer\cite{transformer}         & \textbf{0.437} / \textbf{0.500}  & 2.1 / 2.6 / 4.9 / 13.6 / 45.9 & 13.2 / 31.1 / 90.2 / 310.3 / $\times$ \\
    \multirow{-2}{*}{ORB+Boost-B} & Attention-Free Transformer\cite{aft}  & 0.436 / 0.495  & \textbf{1.6} / \textbf{2.0} / \textbf{3.2} / \textbf{4.3} / \textbf{7.8} & \textbf{8.4} / \textbf{14.5} / \textbf{27.0} / \textbf{51.3} / \textbf{108.2} \\
    \midrule
    % superpoint+boost-f
                                         & Vanilla Transformer\cite{transformer}          & \textbf{0.679} / \textbf{0.777}  & 2.8 / 3.9 / 8.7 / 27.1 / 96.8 & 23.5 / 60.0 / 185.0 / $\times$ / $\times$\\
    \multirow{-2}{*}{SuperPoint+Boost-F} & Attention-Free Transformer\cite{aft}  & 0.669 / 0.758  & \textbf{1.9} / \textbf{2.1} / \textbf{3.3} / \textbf{5.4} / \textbf{10.2} & \textbf{13.2} / \textbf{23.5} / \textbf{44.1} / \textbf{87.3} / \textbf{194.2}\\
    \bottomrule
    % \hline
    \end{tabular}
    }
    \vspace{-0.9em}
    \caption{The efficiency of using different Transformer modules. The table shows the mean matching accuracy (MMA) under thresholds 3 and 5 on HPatches dataset and the runtime for boosting different numbers of local features on RTX 3090 and Jetson Xavier NX 8GB. `$\times$' indicates CUDA running out of memory.
    }
    \label{tab:trans-and-aft}
\end{table*}

\begin{table}[t]
    \centering
    \resizebox{1.\columnwidth}{!}{
    \large
    \begin{tabular}{llcccccc}
    \toprule
    \multicolumn{2}{l}{\multirow{2}{*}{\textbf{Method}}} & \multicolumn{2}{c}{\textbf{Standard}} & \multicolumn{2}{c}{\textbf{Rotated}} & \multicolumn{2}{c}{\textbf{Average}} \\
    \multicolumn{2}{l}{}     & \textbf{@3}       & \textbf{@5}       & \textbf{@3}       & \textbf{@5}      & \textbf{@3}       & \textbf{@5}      \\
    \midrule
    \multirow{4}{*}{SIFT\cite{sift}}       & No boost    & 0.534 & 0.586 & \textbf{0.505} & \textbf{0.559} & \textbf{0.519} & \textbf{0.572} \\
                                & No $\mathcal{L}_{BOOST}$   & 0.571 & \underline{0.644} & 0.216 & 0.236 & 0.393 & 0.440 \\
                                & $\lambda$ = 1  & \textbf{0.577} & \textbf{0.651} & 0.263 & 0.287 & 0.420 & 0.469  \\
                                & $\lambda$ = 10 & \underline{0.573} & 0.640 & \underline{0.391} & \underline{0.428} & \underline{0.482}  & \underline{0.534}  \\
    \midrule
    \multirow{4}{*}{SuperPoint\cite{sp}} & No boost    & 0.654 & 0.738 & 0.202 & 0.222 & 0.428  & 0.480 \\
                                & No $\mathcal{L}_{BOOST}$   & 0.663 & 0.756 & 0.209 & 0.232 & 0.436 & 0.494  \\
                                & $\lambda$ = 1  & \textbf{0.670} & \textbf{0.763} & \textbf{0.218} & \textbf{0.242} & \textbf{0.444} & \textbf{0.503} \\
                                & $\lambda$ = 10 & \underline{0.669} & \underline{0.758} & \underline{0.213} & \underline{0.235} & \underline{0.441}  & \underline{0.497}   \\
    \bottomrule
    \end{tabular}
    }
    \caption{Ablation study on the $\mathcal{L}_{BOOST}$. The table shows the mean matching accuracy (MMA) under thresholds 3 and 5 on the standard HPatches dataset, the rotated HPatches dataset, and the average performance using both datasets. We highlight the \textbf{first} and \underline{second} best MMA values. The result shows the $\mathcal{L}_{BOOST}$ can help the boosted descriptors retain the performance of original descriptors in the cases where the training set does not include. Hence $\mathcal{L}_{BOOST}$ can improve the generalization ability of the trained model.}
    \label{tab:loss-ablation}
\end{table}

We compared the \M~ using different Transformer modules for the cross-boosting stage. Specifically, we present the results of the vanilla transformer using MHA\cite{transformer} and the attention-free transformer using AFT\cite{aft} in \cref{tab:trans-and-aft}. The results show that the Attention-Free Transformer is much faster and consumes less GPU memory than the vanilla one, with a minor drop in matching performance.

\section{Ablation study of the training loss}
\label{sec:loss-abla}
In this section, we study the impact of the training loss on our \M. 
Our training loss consists of two term: $\mathcal{L}_{AP}$ and $\mathcal{L}_{BOOST}$, which are balanced using a weight $\lambda$. 
We use the HPatches\cite{hpatches} for the ablation study following the way in Section 5.1. 
To further evaluate the importance of $\mathcal{L}_{BOOST}$, we additionally use the rotated HPatches dataset\cite{rord} by applying random in-plane rotation of images from 0$^{\circ}$ to 360$^{\circ}$, while our training set MegaDepth\cite{megadepth} does not contain large in-plane rotation cases.

\cref{tab:loss-ablation} shows MMA (Mean Matching Accuracy) results under re-projection error thresholds of $3$ and $5$ pixels for three settings: standard, rotated, and average, which means using the standard HPatches dataset, the rotated HPatches dataset, and the average performance of using both datasets respectively.
We can see that the original SIFT\cite{sift} achieves the best result under the rotated HPatches in the rotated and average settings. 
We believe the reason is that the training set (MegaDepth\cite{megadepth}) does not contain large-in-plane rotation cases. 
However, our $\mathcal{L}_{BOOST}$ can help the boosted SIFT retain the performance of SIFT on rotated HPatches when $\lambda$ increases. 

We also can see that the boosted SIFT and SuperPoint\cite{sp} can achieve better performance on Standard HPatches when $\lambda=1$, but we set $\lambda=10$ in the paper for a greater generalization of our method.

\section{Threshold for ratio/distance test}
\label{sec:threshold}
It is known that using ratio or distance tests can reject many incorrect correspondences and improve the RANSAC\cite{ransac} efficiency and the final matching results. The ratio test is to check if the ratio of the descriptor distance of the closest feature to that of the second closest one is smaller than a threshold. Distance tests simply check if the distance between two matched descriptors is within a threshold.

To find a suitable ratio/distance threshold for a fair comparison in the experiments, we compute the probability density functions (PDFs) of correct and incorrect matches following \cite{sift} and select thresholds for all descriptors according to the threshold criteria of their corresponding baselines.
We use HPatches dataset\cite{hpatches} to compute the PDFs like D2-Net\cite{d2net}. 
The PDFs are shown in \cref{fig:pdfs}.

We use ratio tests for matching DoG-based descriptors (\eg RootSIFT\cite{rootsift}, SOSNet\cite{sosnet} and our boosted SIFTs) like SIFT\cite{sift}. Specifically, we adopt Lowe's recommended threshold of 0.8\cite{sift} for SIFT, RootSIFT and SIFT+Boost-B, while for SOSNet and SIFT+Boost-F we use a ratio threshold of 0.85.

We use distance tests instead of ratio tests for matching ORB\cite{orb} and ORB+Boosted-B descriptors since ratio tests do not work well for those descriptors. The selected distance thresholds are 45 and 50 respectively. 

We use distance tests for matching SuperPoint\cite{sp} descriptors and use the same distance threshold of 0.7 as for HLoc\cite{hloc}. We select 0.8 and 55 as the distance thresholds for matching SuperPoint+Boost-F and SuperPoint+Boost-B descriptors respectively.

Regarding the ALIKE-based descriptor, the distinctions between correct and incorrect matches in the PDF curves are unclear. We heuristically use a ratio threshold of 0.9 for both ALIKE\cite{alike} and our ALIKE+Boost-F, and a threshold of 0.88 for our ALIKE+Boost-B, which can retain 77.3\%/77.6\%/77.4\% correct matches while filtering out 94.4\%/94.6\%/91.3\% incorrect matches.

\section{Qualitative examples}
\label{sec:example}
\cref{fig:aachen} and \cref{fig:inloc} show some matching results using different descriptors on Aachen Day-Night v1.1\cite{aachenv1.1} and InLoc\cite{inloc}.

\begin{figure*}[h]
\centering
\begin{subfigure}{0.32\textwidth}
    \includegraphics[width=\textwidth]{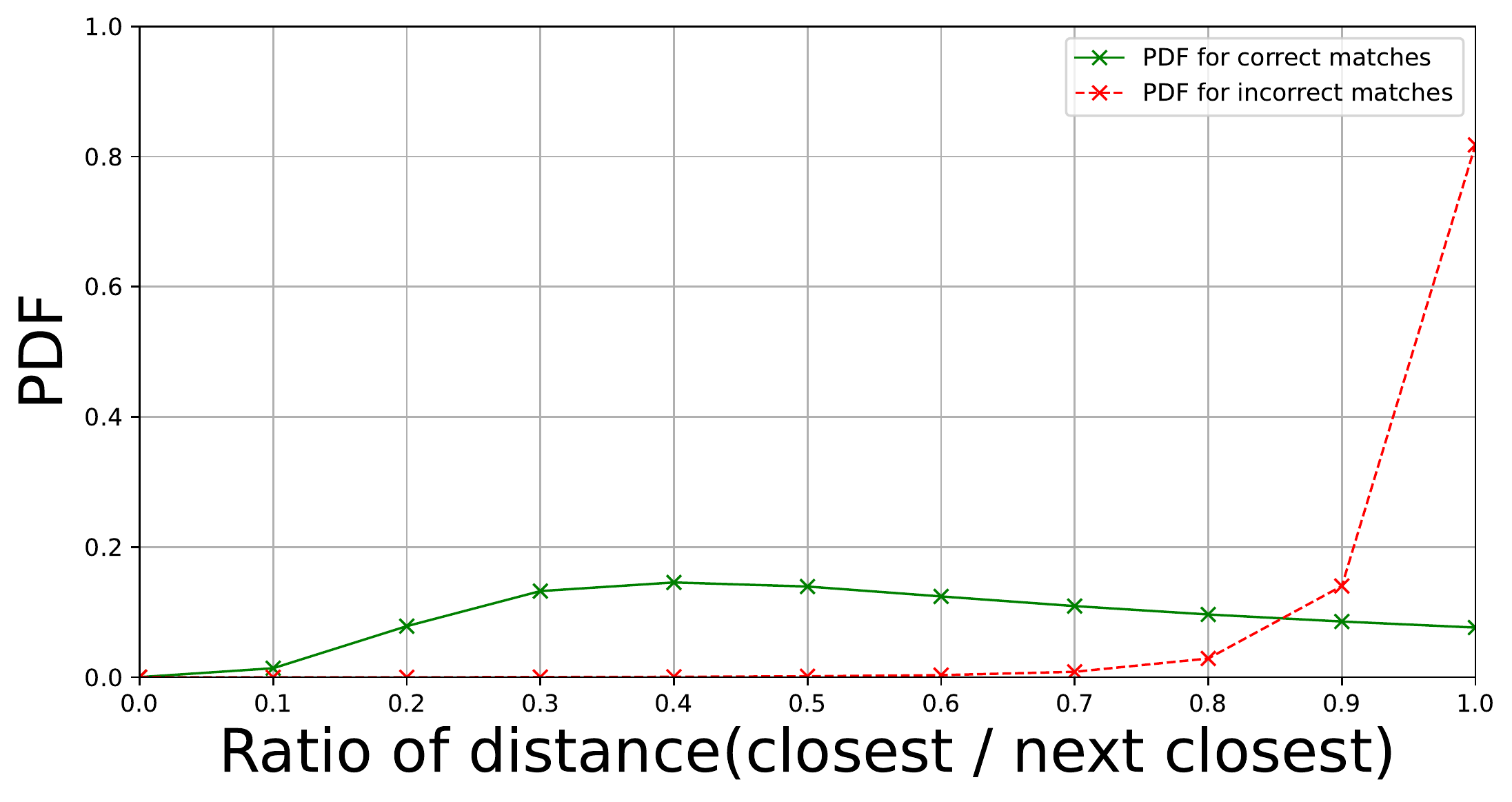}
    \caption{SIFT}
    \label{fig:sift}
\end{subfigure}
\hfill
\begin{subfigure}{0.32\textwidth}
    \includegraphics[width=\textwidth]{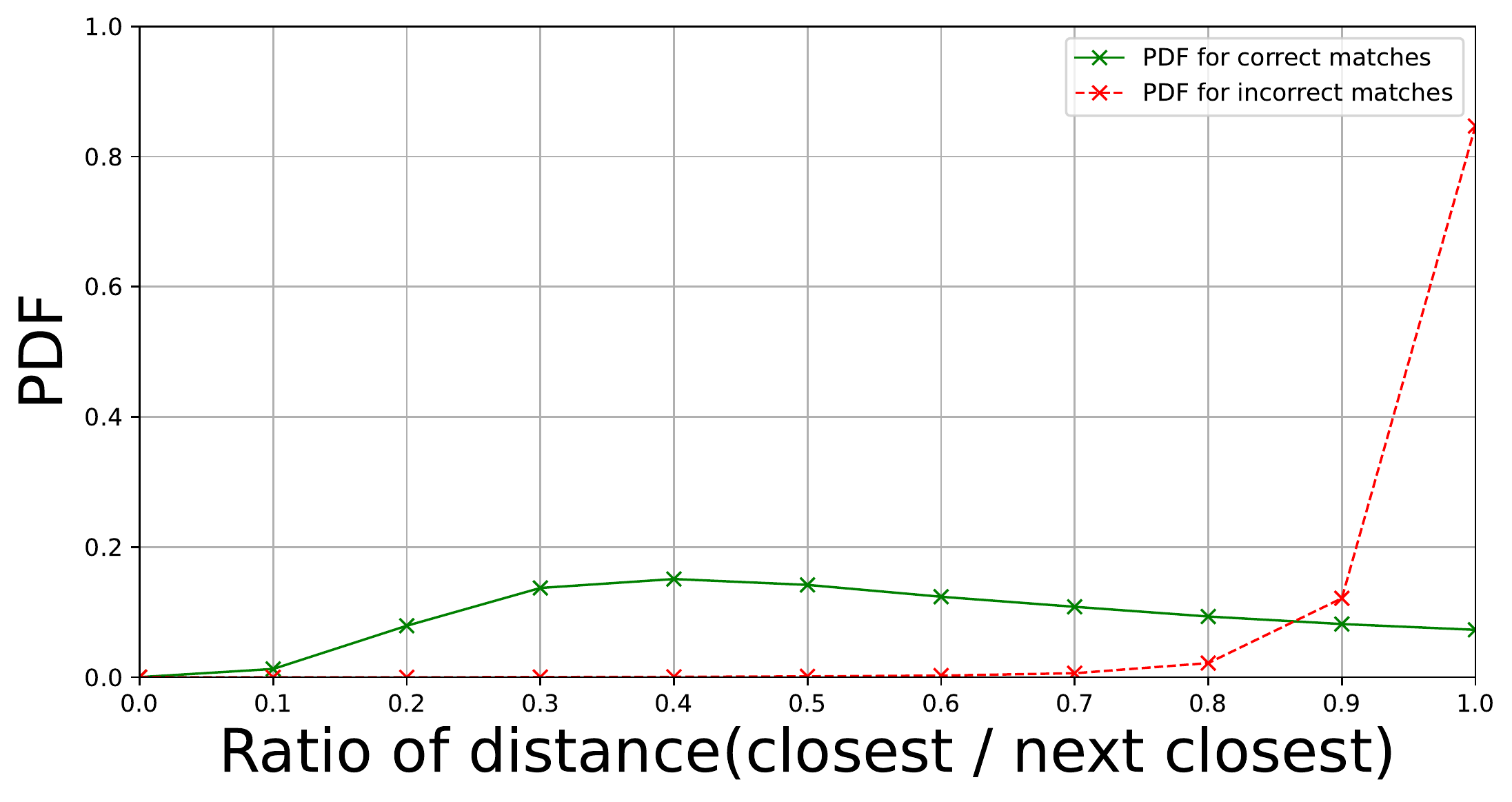}
    \caption{RootSIFT}
    \label{fig:rootsift}
\end{subfigure}
\hfill
\begin{subfigure}{0.32\textwidth}
    \includegraphics[width=\textwidth]{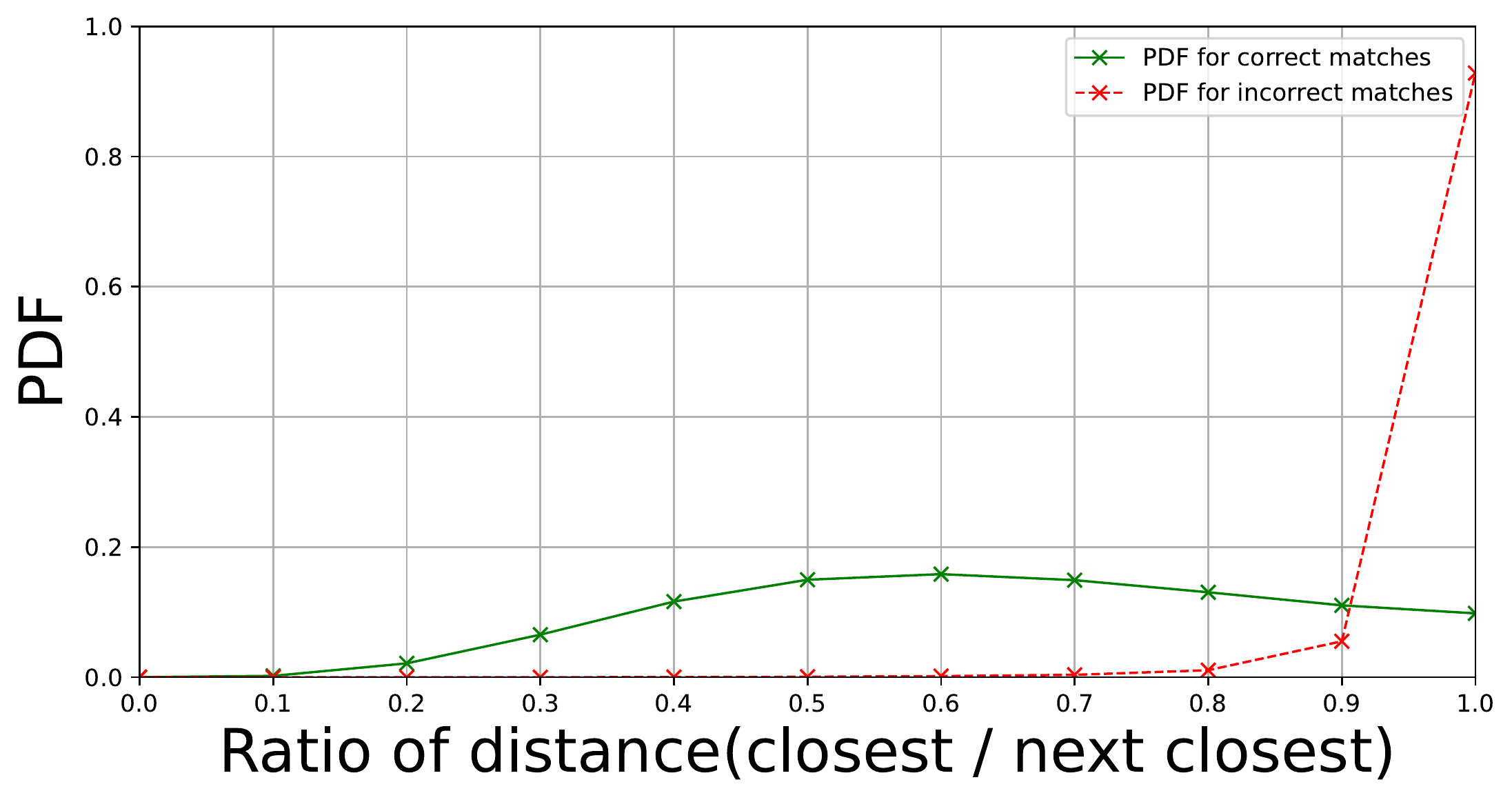}
    \caption{SOSNet}
    \label{fig:sosnet}
\end{subfigure}
\begin{subfigure}{0.32\textwidth}
    \includegraphics[width=\textwidth]{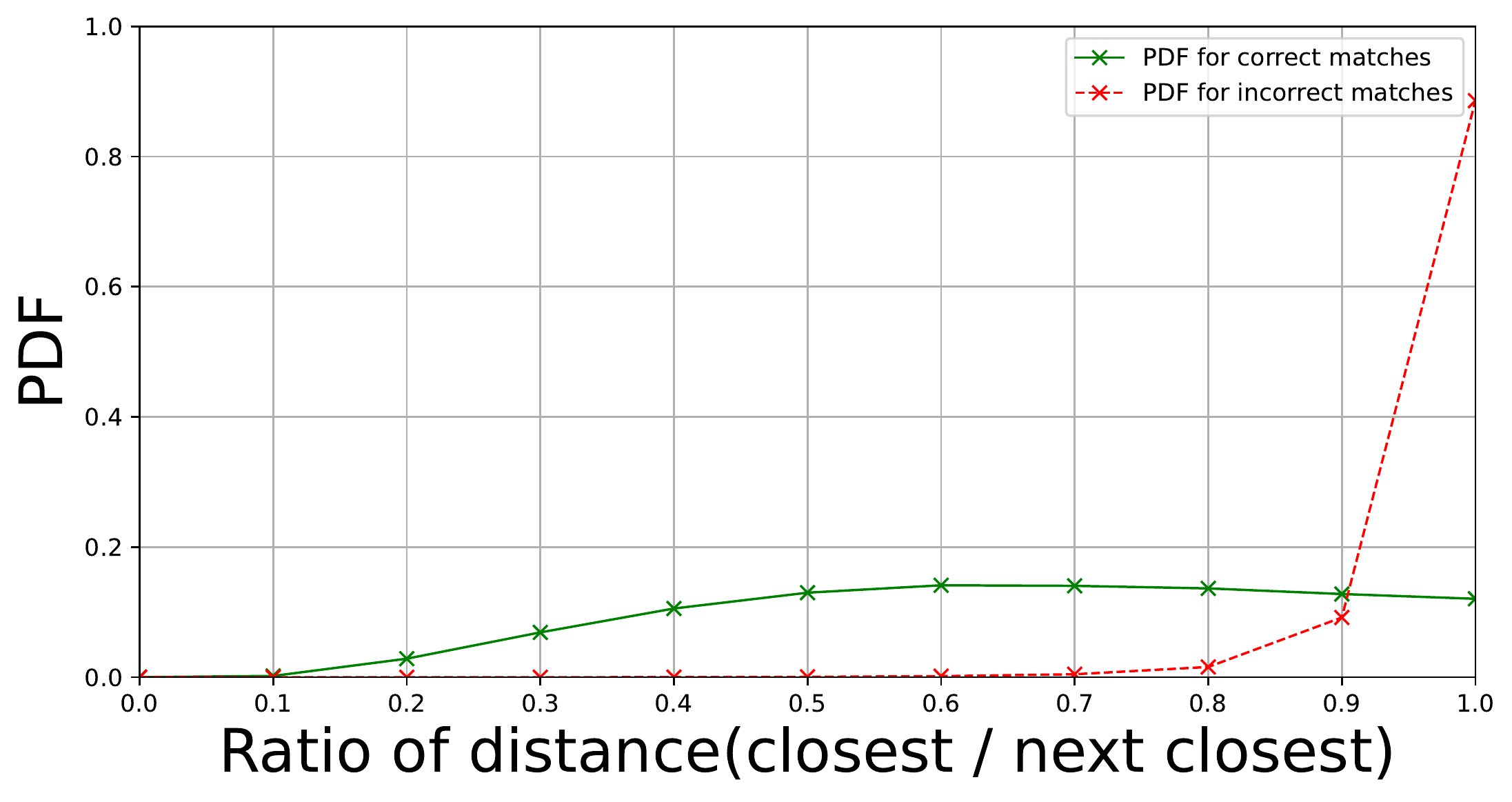}
    \caption{SIFT+Boost-F}
    \label{fig:sift-boost-f}
\end{subfigure}
\hfill
\begin{subfigure}{0.32\textwidth}
    \includegraphics[width=\textwidth]{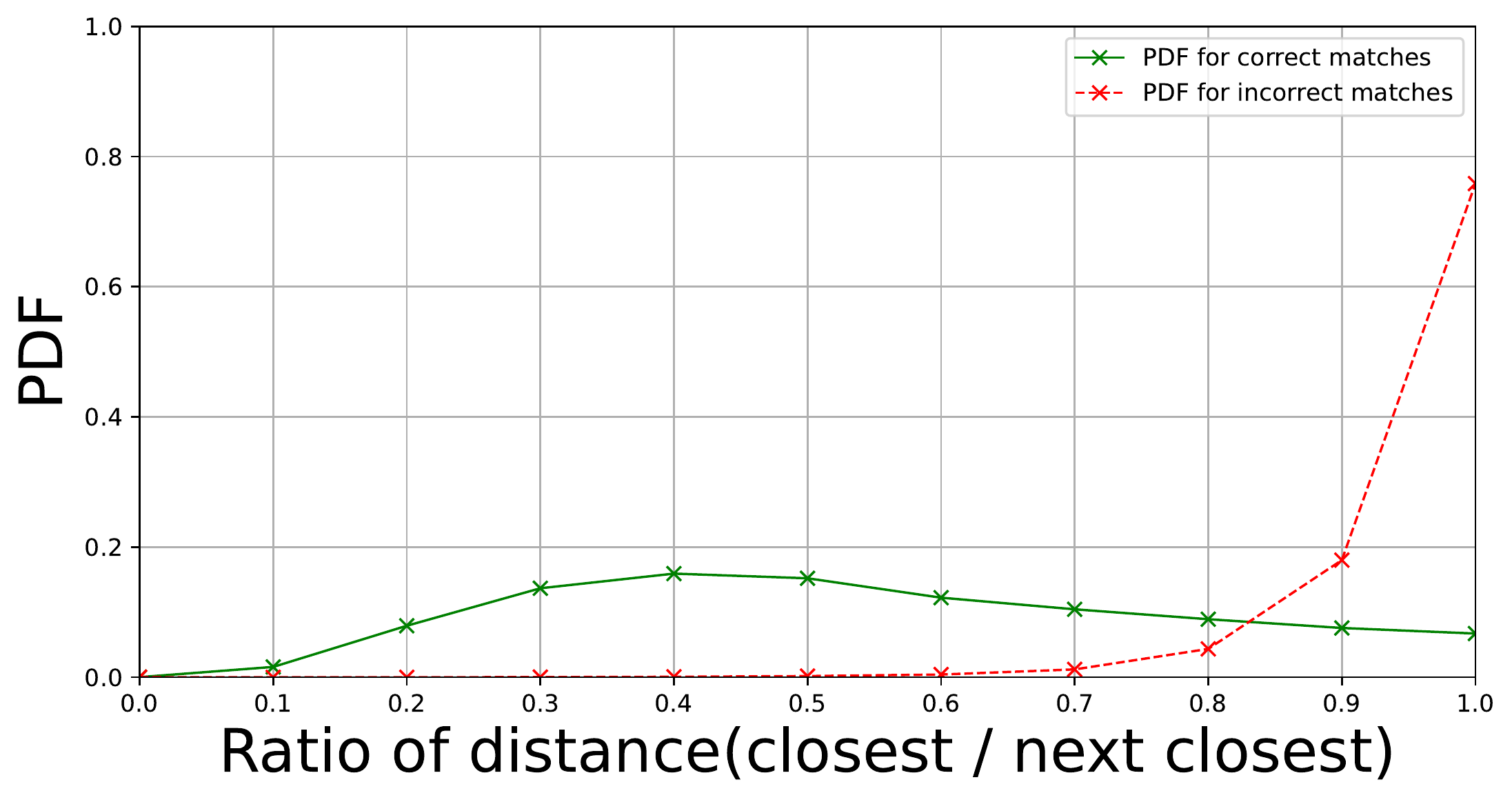}
    \caption{SIFT+Boost-B}
    \label{fig:sift-boost-b}
\end{subfigure}
\hfill
\begin{subfigure}{0.32\textwidth}
    \includegraphics[width=\textwidth]{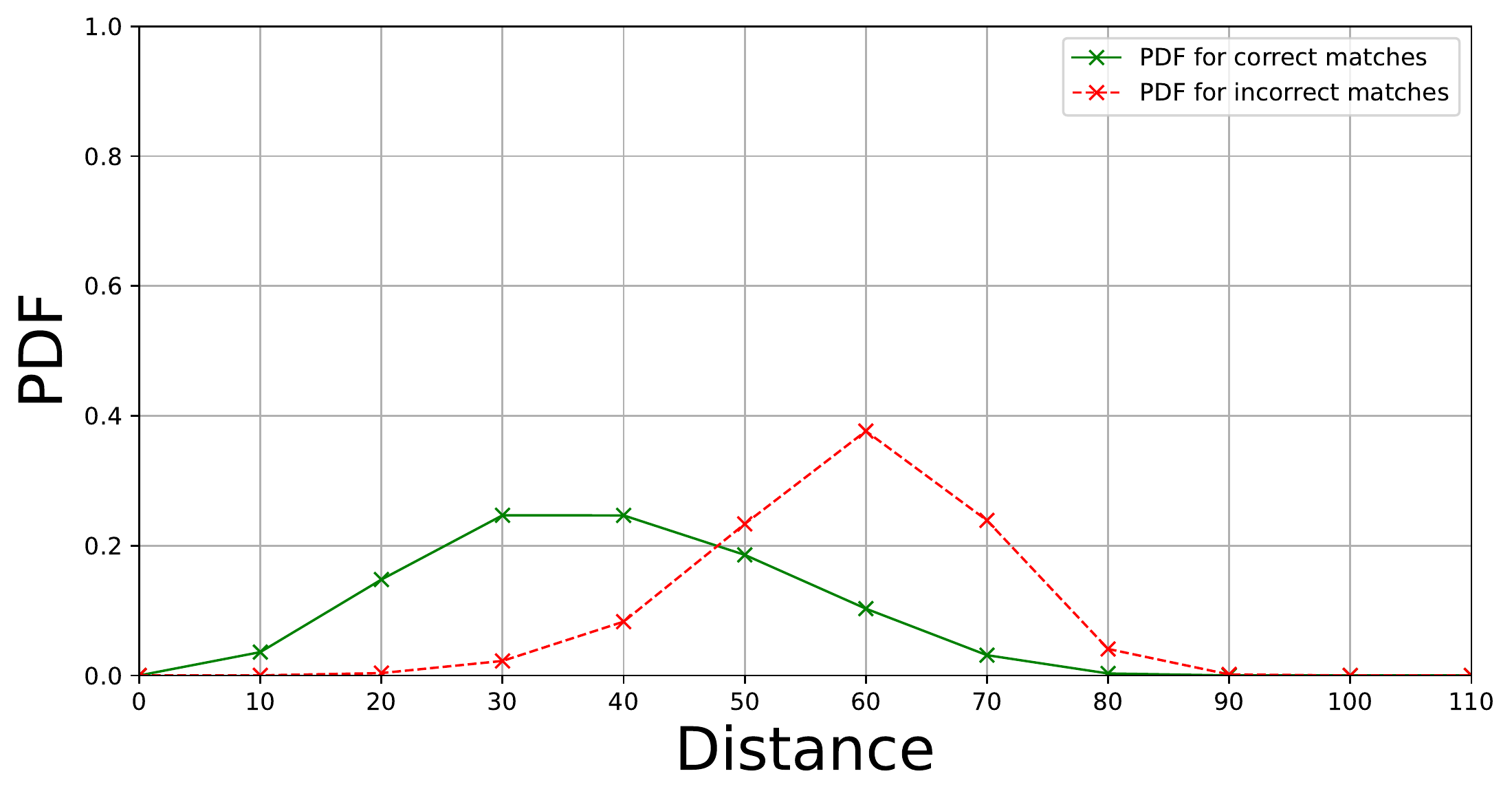}
    \caption{ORB}
    \label{fig:orb}
\end{subfigure}
\begin{subfigure}{0.32\textwidth}
    \includegraphics[width=\textwidth]{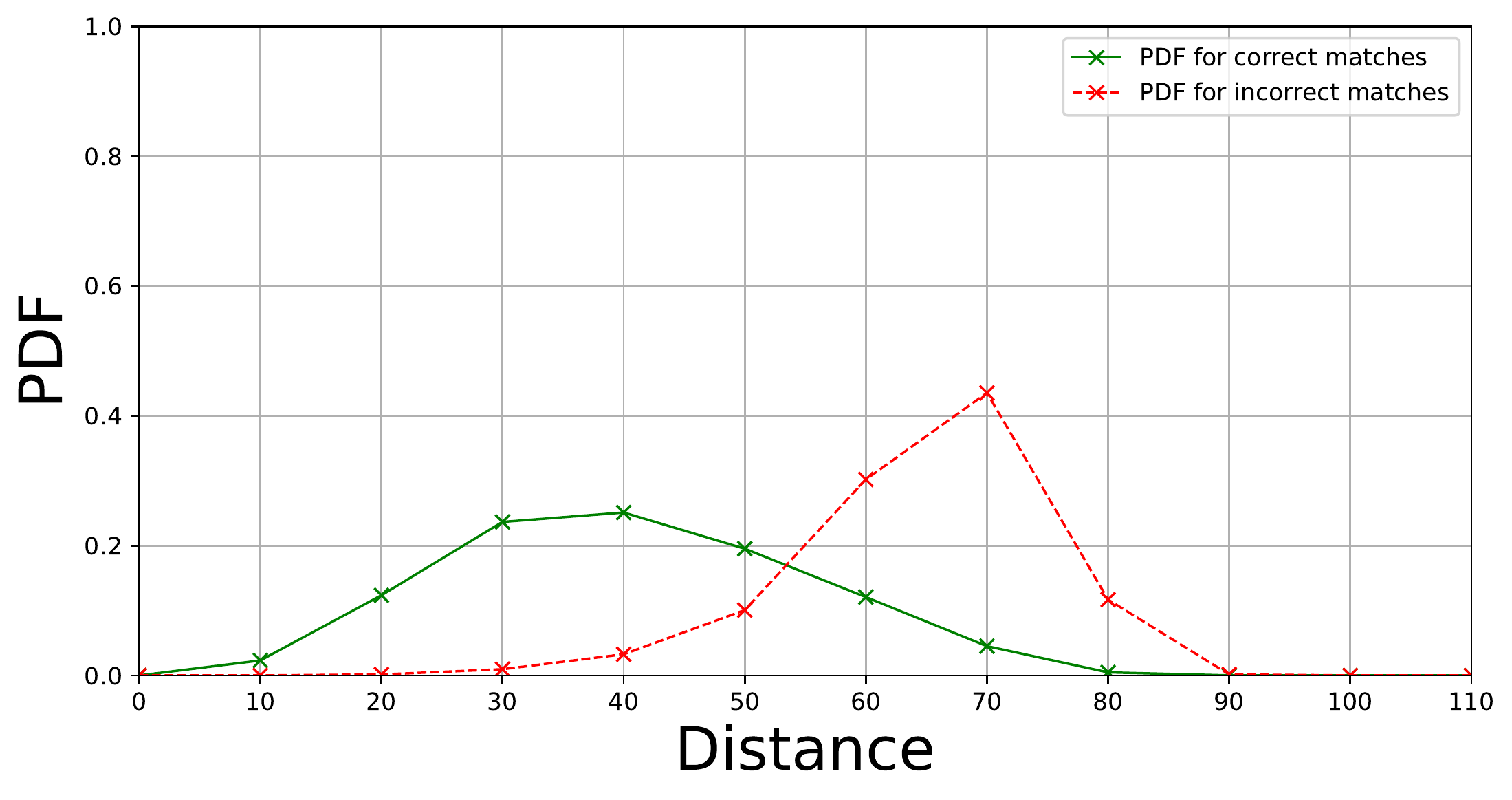}
    \caption{ORB+Boost-B}
    \label{fig:orb-boost-b}
\end{subfigure}
\hfill
\begin{subfigure}{0.32\textwidth}
    \includegraphics[width=\textwidth]{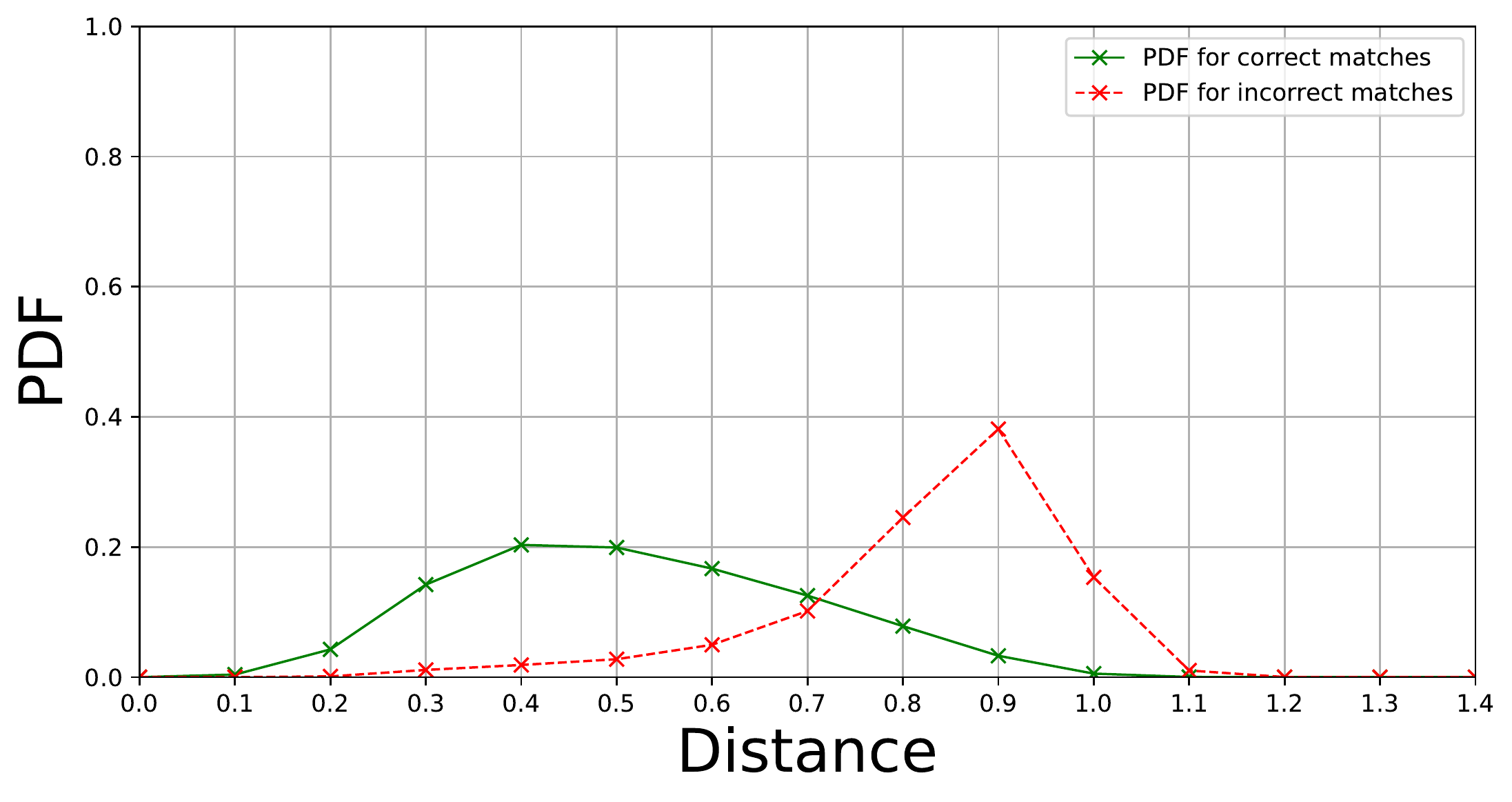}
    \caption{SuperPoint}
    \label{fig:sp}
\end{subfigure}
\hfill
\begin{subfigure}{0.32\textwidth}
    \includegraphics[width=\textwidth]{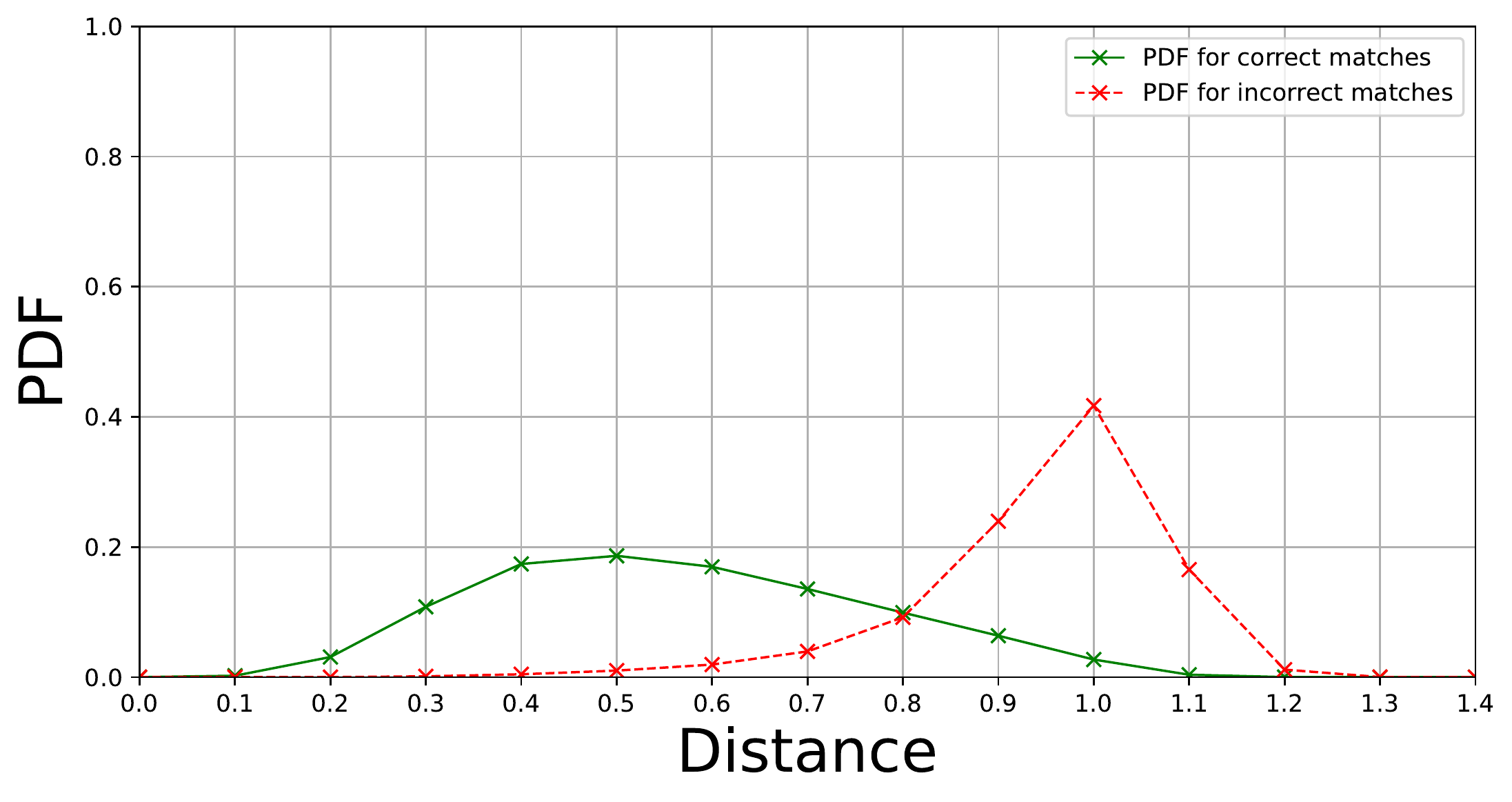}
    \caption{SuperPoint+Boost-F}
    \label{fig:sp-boost-f}
\end{subfigure}
\begin{subfigure}{0.32\textwidth}
    \includegraphics[width=\textwidth]{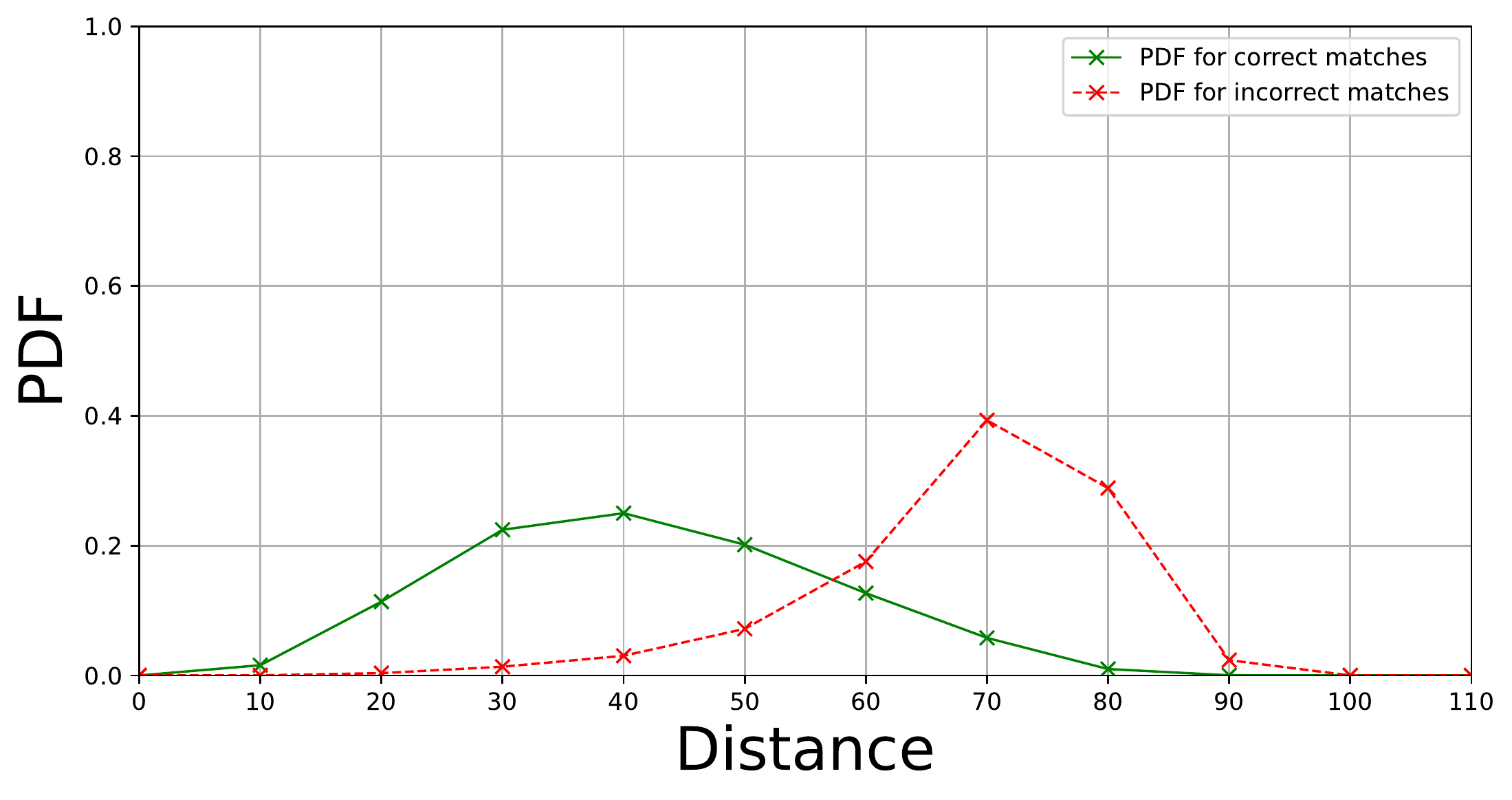}
    \caption{SuperPoint+Boost-B}
    \label{fig:sp-boost-b}
\end{subfigure}
\hfill
\begin{subfigure}{0.32\textwidth}
    \includegraphics[width=\textwidth]{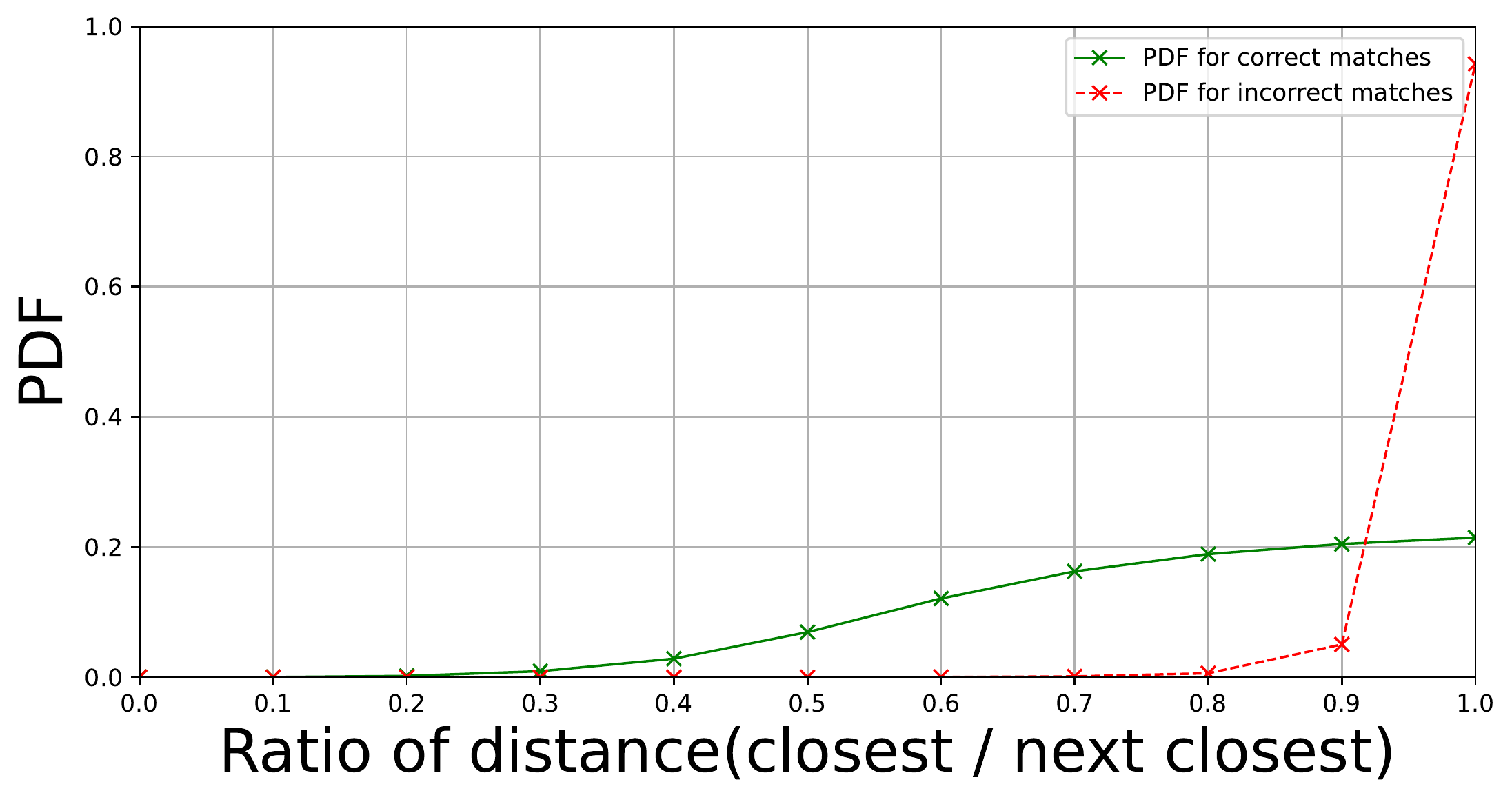}
    \caption{ALIKE}
    \label{fig:alike}
\end{subfigure}
\hfill
\begin{subfigure}{0.32\textwidth}
    \includegraphics[width=\textwidth]{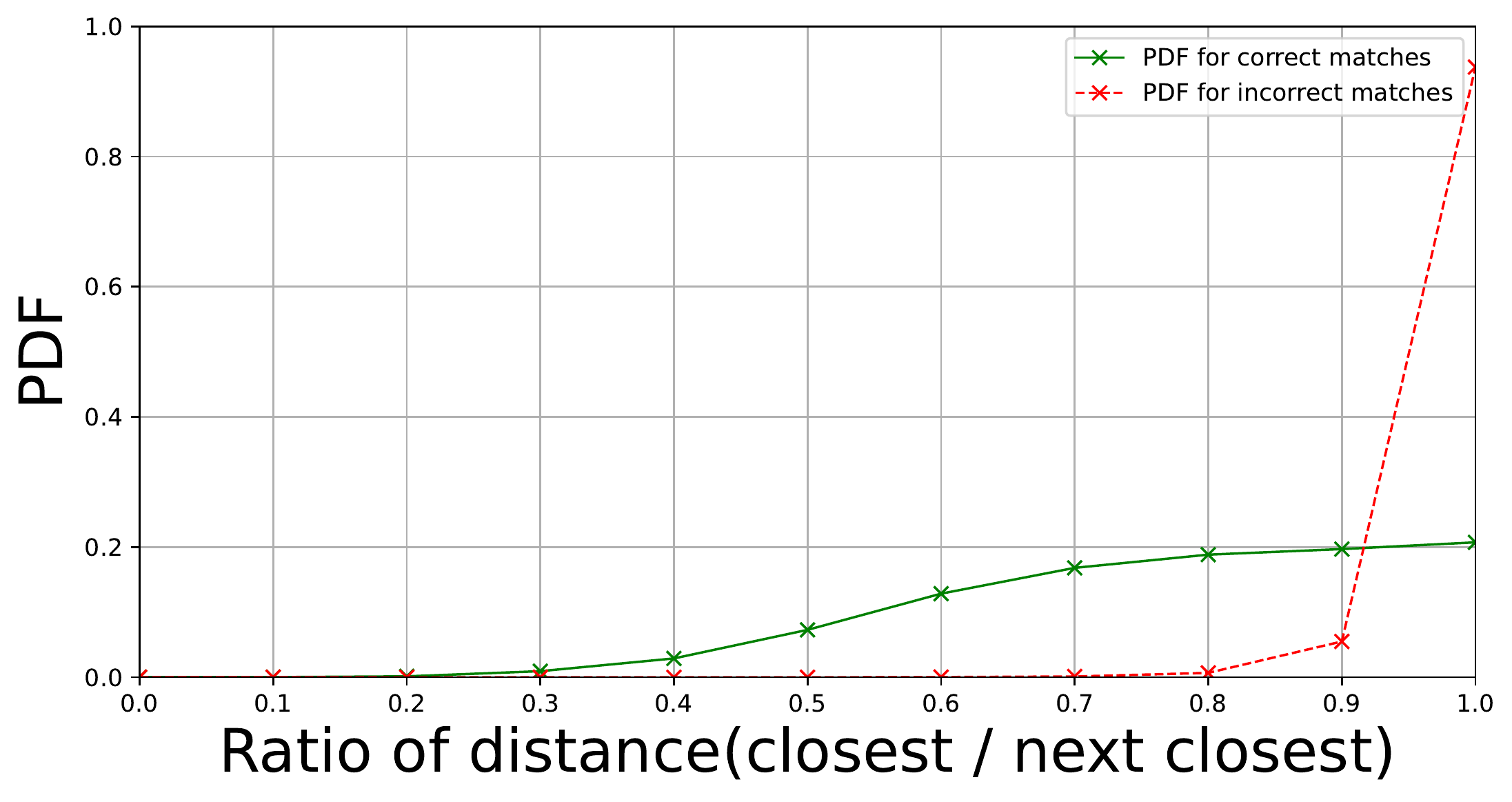}
    \caption{ALIKE+Boost-F}
    \label{fig:alike-boost-f}
\end{subfigure}
\begin{subfigure}{0.32\textwidth}
    \includegraphics[width=\textwidth]{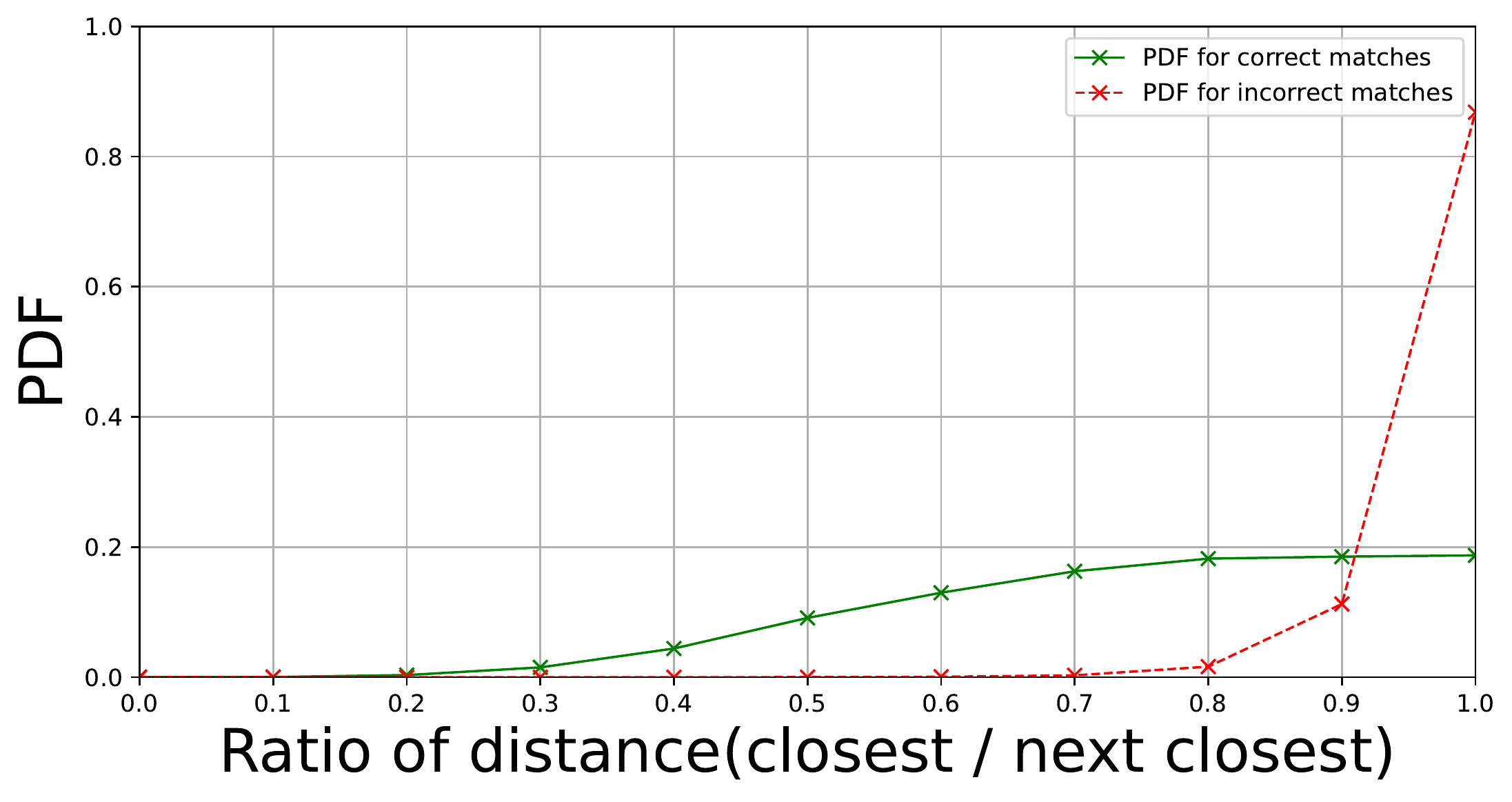}
    \caption{ALIKE+Boost-B}
    \label{fig:alike-boost-b}
\end{subfigure}
\caption{Ratio or distance PDFs for different descriptors. We compute the PDFs for all the descriptors using HPatches dataset\cite{hpatches}. For correct matches, the distance between the warp points and the keypoints is below 4 pixels. For incorrect matches, the distance is greater than 10 pixels.}
\label{fig:pdfs}
\end{figure*}

\begin{figure*}[ht]
  \centering
%   \fbox{\rule{0pt}{2in} \rule{0.9\linewidth}{0pt}}
  \includegraphics[width=1.0\linewidth]{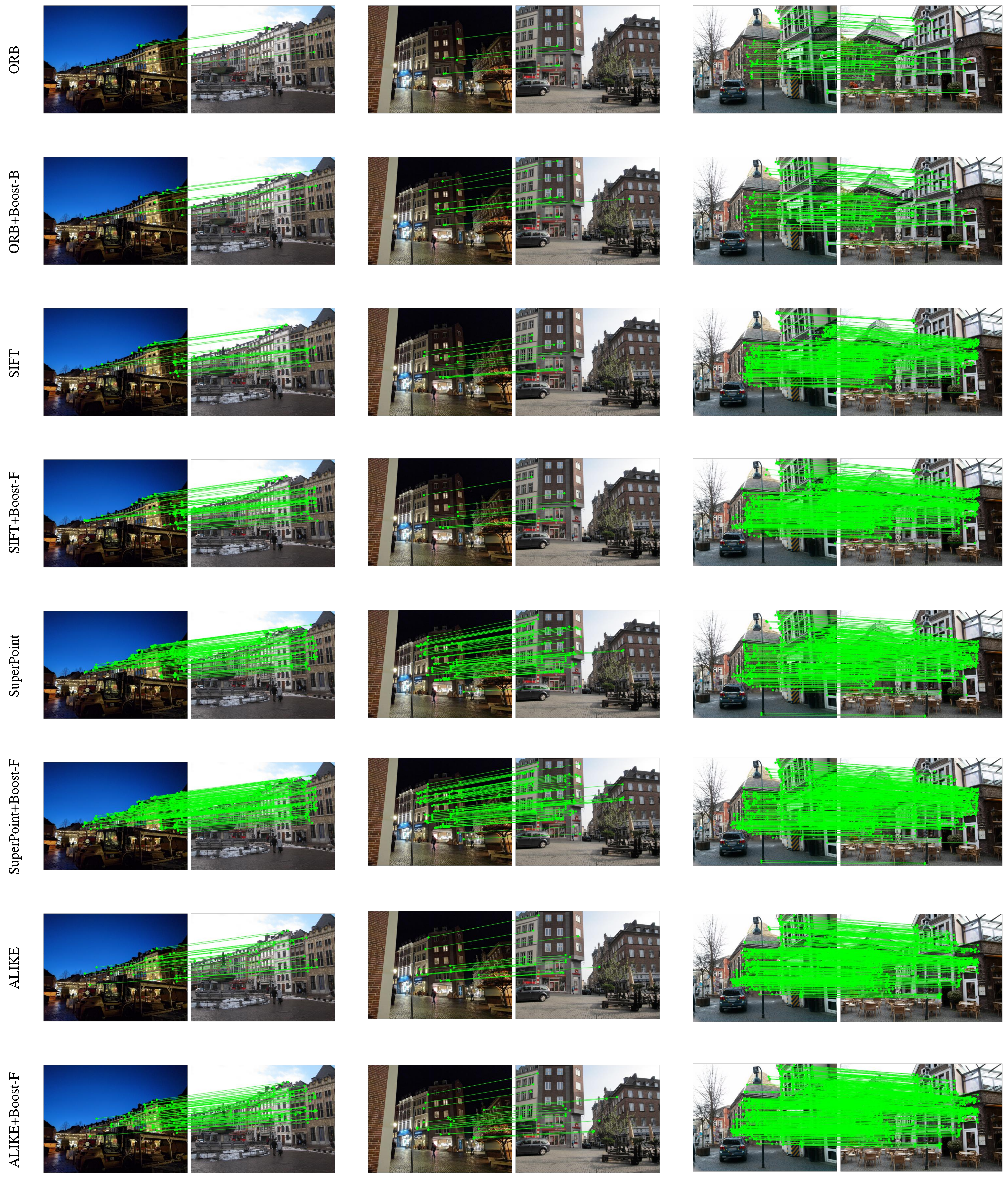}
  \caption{Matching results of using different descriptors on Aachen Day-Night v1.1\cite{aachenv1.1}. By boosting the original descriptors, our methods (represented by 'xxx+Boost-x') can produce more correct matches under significant changes in viewpoint and illumination. More results on InLoc \cite{inloc} are shown in \cref{fig:inloc}.}
  \label{fig:aachen}
\end{figure*}

\begin{figure*}[ht]
  \centering
%   \fbox{\rule{0pt}{2in} \rule{0.9\linewidth}{0pt}}
  \includegraphics[width=0.8\linewidth]{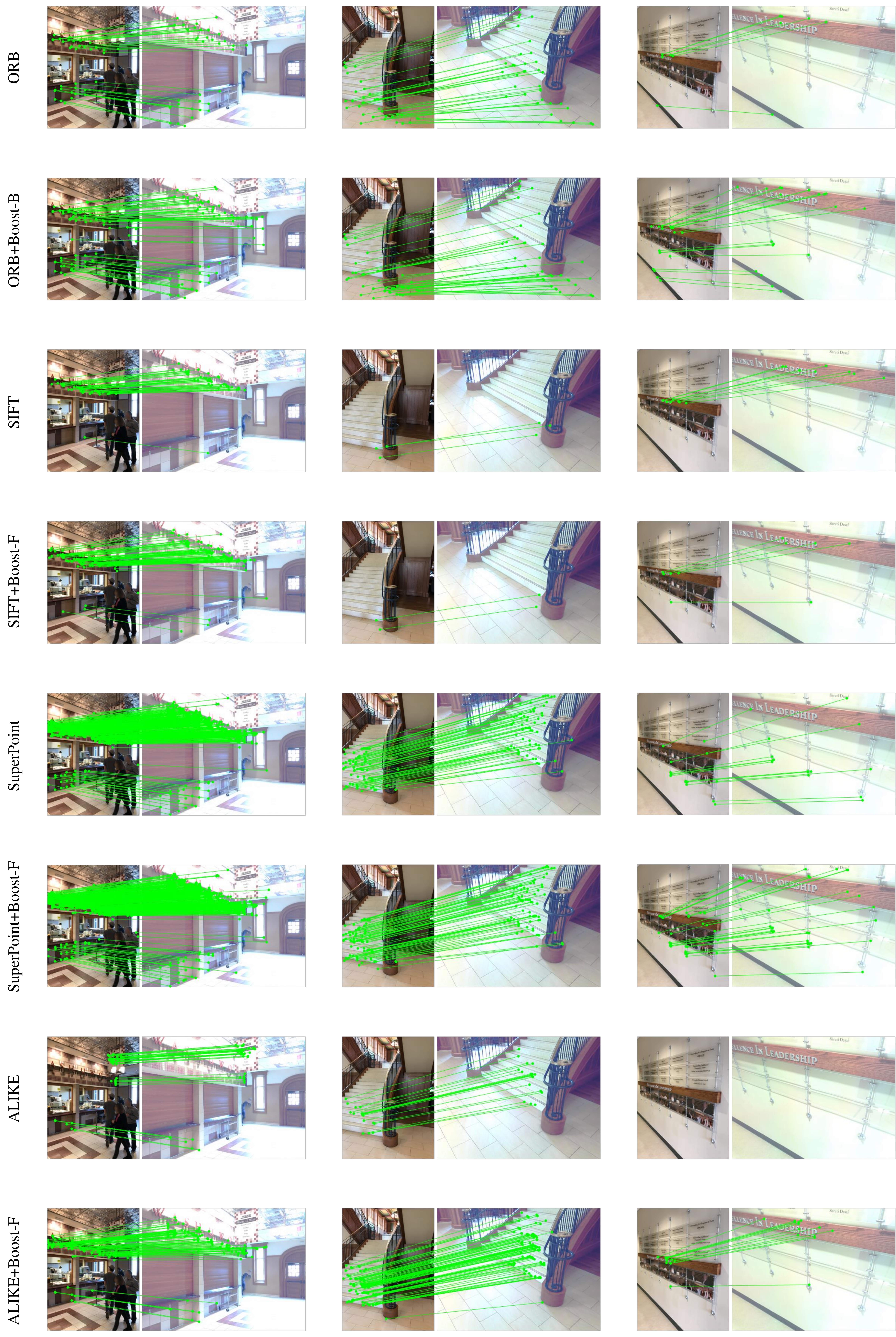}
  \caption{Matching results on InLoc dataset\cite{inloc}. Our methods can boost the performance of descriptors under significant changes in viewpoint and texture-less areas. We also can see the failure case, where SIFT even performs worse than ORB and SIFT+Boost-F cannot improve the performance in those indoor scenes.}
  \label{fig:inloc}
\end{figure*} \clearpage \clearpage \fi

%%%%%%%%% REFERENCES
{\small
\bibliographystyle{ieee_fullname}
\bibliography{egbib}
}

\end{document}